%% file: paper.tex
\documentclass[]{fairmeta}

\usepackage{amsmath,amssymb}

\newtcolorbox{finding}[1]{%
  breakable, enhanced,
  colback=black!5, colframe=black!45,
  boxrule=0.6pt, arc=4pt,
  left=8pt, right=8pt, top=6pt, bottom=6pt,
  before skip=8pt, after skip=8pt,
  before upper={\bfseries\itshape Finding~#1:\quad\normalfont}%
}

\definecolor{skyblue}{RGB}{30,160,235}

\definecolor{manlingteal}{RGB}{0,135,120}

\title{HumanCLAW: Can Vision-Language Models Act Through a Body? }

\author[1,2,*,\ddag]{\mbox{Li Siyao}}
\author[3,*,]{\mbox{Jiawei Gu}}
\author[2,\#]{\mbox{Shuai Liu}}
\author[2,\#]{\mbox{Kairui Hu}}
\author[1,4,\ddag]{\mbox{Zekun Li}}
\author[3]{\mbox{Linjie Li}}
\author[1]{\mbox{Chengcheng Tang}}
\author[1]{\mbox{Po-Chen Wu}}
\author[1]{\mbox{Ivan Shugurov}}
\author[1]{\mbox{Lingni Ma}}
\author[1]{\mbox{Michael Zollhoefer}}
\author[1]{\mbox{Sizhe An}}
\author[1,\dag]{\mbox{Abhay Mittal}}
\author[1,\dag]{\mbox{Amy Zhao}}
\author[3,\dag]{\mbox{Ranjay Krishna}}
\author[5,\dag]{\mbox{Manling Li}}
\author[2,\dag]{\mbox{Ziwei Liu}}
\author[1,\dag]{\mbox{Chuan Guo}}

\affiliation[1]{Meta}
\affiliation[2]{Nanyang Technological University}
\affiliation[3]{University of Washington}
\affiliation[4]{Brown University}
\affiliation[5]{Northwestern University}

\contribution[*]{Equal contribution}
\contribution[\ddag]{Intern at Meta}
\contribution[\#]{Equal contribution}
\contribution[\dag]{Equal advice}
\input{content/0_abstract}

\metadata[Project Page]{\url{https://human-claw.github.io}}
\date{\today}

\begin{document}

\maketitle

\input{content/1.9_intro}
\input{content/3_method}

\input{content/4_hcb}
\input{content/5_experiment}

\input{content/2_related}

\input{content/6_discussion}

\clearpage
\newpage
\bibliographystyle{assets/plainnat}
\bibliography{paper}

\clearpage
\newpage
\beginappendix

\input{content/appendix}

\end{document}

%% file: content/0_abstract.tex
\abstract{%
Evaluating whether a vision-language model (VLM) can act through a physical
body is challenging. The outcome of an action couples the VLM's decision
with motor control. When a task fails, it is hard to tell whether the VLM
made a bad choice or the motor controller simply failed to execute it, e.g.,
losing balance and falling. In this work, we introduce
\textbf{HumanCLAW}, an evaluation framework that decouples action
decision-making from low-level execution. At every step, a harnessed,
off-the-shelf VLM issues an atomic skill command, and the command is
translated into a sub-second chunk of continuous full-body motion with real physical
consequences, including gravity and collisions. The body can therefore act
freely in the physical world, while execution-side disturbances, balance and
motor errors, are factored out. What remains measurable is the model's
\textbf{action intelligence}: its moment-to-moment choice of what the body
should execute next. Based on this framework, we build \textbf{HumanCLAW-Bench}:
$1{,}218$ long-horizon, egocentric \texttt{find-navigate-interact} episodes
across $41$ indoor scenes. We test nine
state-of-the-art VLMs and find that none solves the benchmark; the best
model reaches only a $16.8\%$ success rate. Recognizing the target is not
the bottleneck. What current VLMs lack is \textbf{embodied self-awareness}:
they lose track of their own body, failing to tell where it is, whether it
has reached the goal, or whether it has hit an obstacle.%
}

%% file: content/1.9_intro.tex
\section{Introduction}
\label{sec:introduction}

\begin{figure*}[t]
  \centering
  \includegraphics[width=0.98\linewidth]{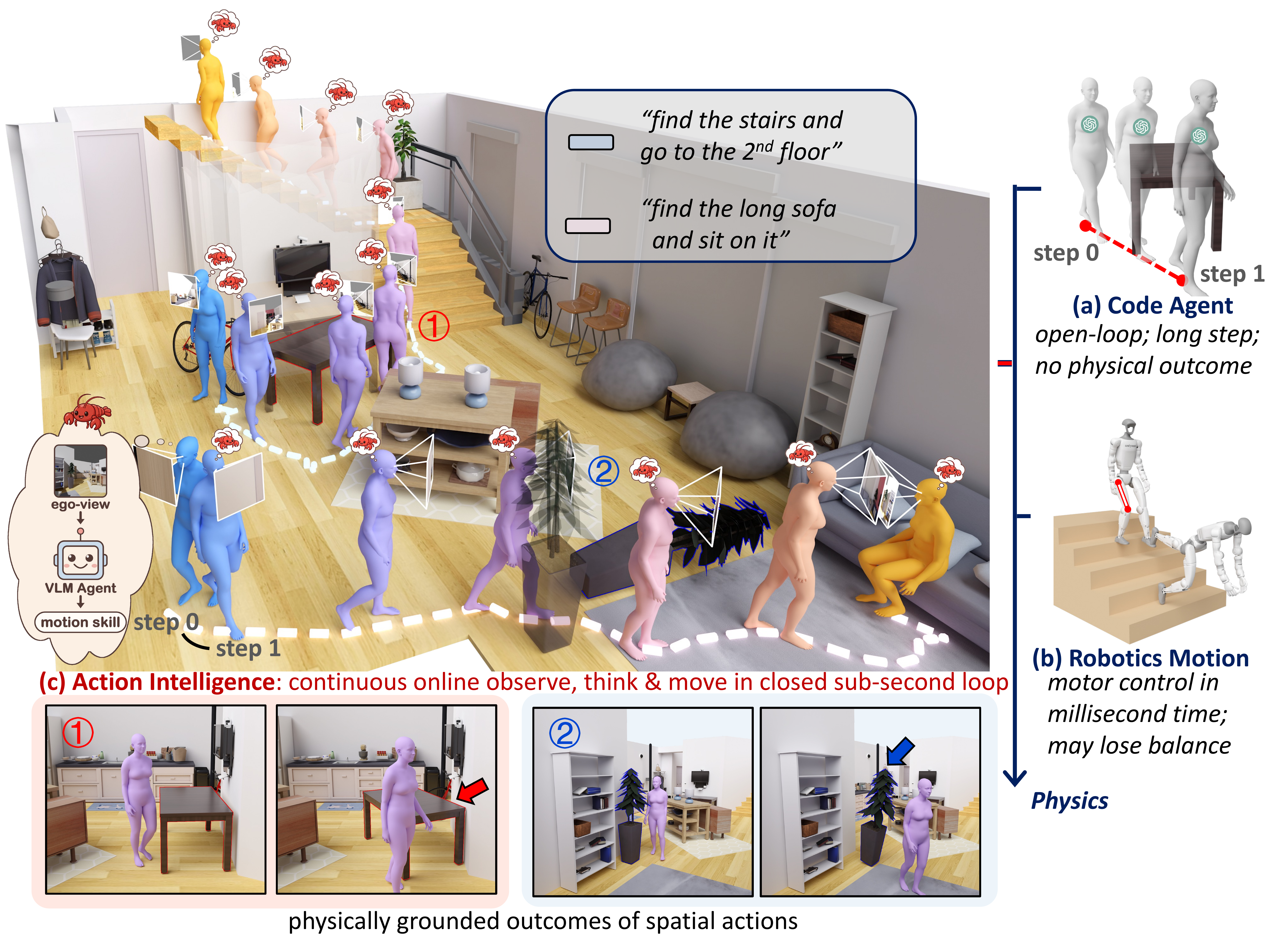}
  \caption{\textbf{Where does action intelligence live?}  Three regimes of action
  along increasing physical embodiment: \textbf{(a)} action without a body, and \textbf{(b)} a body
  whose action is inseparable from balance and motor control. \textbf{(c)} HumanCLAW
  keeps body and world in a closed sub-second loop but factors out low-level control,
  so a failed episode reads as a wrong decision, not a lost balance. \textbf{Main
  panel:} one find-navigate-interact episode, each posed body a sub-second decision.
  \textbf{Bottom:} the physical outcomes it produces.}
  \label{fig:teaser}
\end{figure*}

Recent vision-language models (VLMs) achieve strong performance on tasks
involving visual grounding, spatial relations, and spatial
reasoning~\citep{chen2024spatialvlm,yang2024thinking}.
If such a model is given a physical body, an egocentric observation, a
history of its actions, and a goal, can it determine an \textit{executable
action} appropriate for the current moment?
A positive answer would be tantalizing: the decision would come from
general-purpose reasoning rather than from a policy fitted to a specific
embodiment trajectory; an agent built on this reasoning could act in
scenarios it was never trained on, without collecting massive action
demonstrations for the decision maker.
We call this capacity \textbf{action intelligence}: the ability to decide,
moment by moment, what a physical body should do next inside the
\emph{execution loop}, where each decision is executed and its physical
outcome informs the next.
We view it as the operational component of \textit{spatial intelligence},
converting spatial understanding into closed-loop embodied decisions by
selecting, parameterizing, and sequencing actions as their physical
consequences unfold.

In this paper, we explore whether recent VLMs have action intelligence.
Yet turning a VLM's decision into executable motion is not trivial.
For a simple embodiment, the path is short: a $7$-DoF end-effector action can
be directly written out as text~\citep{yang2025embodiedbench}.
A complex embodiment like the human body offers no shortcut.
With well over a hundred pose dimensions changing many times per second,
full-body motion is far too much to write out as text at every control step.
Existing evaluations largely fall on either side of this gap between text
command and executable motion.
On one side, agent benchmarks abstract away the motion: to keep
evaluation tractable, commands are executed through simulator-defined action
abstractions, scripts, or pre-animated
routines~\citep{puig2018virtualhome,shridhar2020alfred,liu2024visualagentbench,chang2025partnr},
so the model never has to realize full-body behavior whose physical
consequences emerge from the motion itself.
On the other, end-to-end VLAs do learn to generate executable
motion~\citep{zitkovich2023rt2,kim2025openvla,black2024pi0,lee2025molmoact},
but fitting on embodiment-specific trajectories entangles the decision with
the learned control policy, making it difficult to isolate how far the
generalist's reasoning itself extrapolates into physical action.
Moreover, for a humanoid, motor execution can further confound evaluation:
the body may, for example, lose balance while climbing stairs, producing a
failure unrelated to the action decision.
Therefore, measuring action intelligence requires what neither side provides:
a setting where the decision is \textbf{decoupled} from execution, yet the body
still \emph{acts freely} in the physical world (\Cref{fig:teaser}).

To this end, we first endow an off-the-shelf VLM with the ability to act freely
in the physical world.
\textbf{HumanCLAW} closes this gap without training the decision maker.
At each sub-second step, the harnessed VLM takes as input its egocentric view
and text history, reasons explicitly about what it sees, where its body stands,
and what to do next, and outputs one \emph{atomic} whole-body skill with
parameters, checked by a skill-specific verifier.
Atomic is the operative word: a skill is a minimal, unambiguous unit of bodily
action---walk at a given stride, turn by a given angle, sit at a given
height---never a compositional instruction like \textit{``sit on the sofa,''}
which silently bundles finding, approaching, orienting, and committing into
one fuzzy command.
Keeping the vocabulary atomic pushes this composition into the VLM's own
step-by-step reasoning, with no task-specific action training for the decision
maker.
A skill-conditioned motion generator supplies what the VLM cannot emit: it
realizes each accepted skill as continuous, hundred-DoF human motion.
A \textbf{half-physics} simulator then applies its physical consequences
through contact, collision, gravity, and object displacement, while factoring
out balance and motor-tracking failures.
The resulting egocentric view returns for the next decision, closing the loop
(\Cref{fig:pipeline}).
In effect, the VLM is playing a first-person
game~\citep{tan2024cradle,zhang2025videogamebench}, but through a richer,
physically grounded abstraction: the action space consists of parameterized whole-body skills rather
than key presses, and outcomes are determined by physical interaction rather
than by script.
Each step thus pairs an explicit decision with its physical outcome, allowing
failures to be attributed at the skill-decision level.

Based on HumanCLAW, we then propose \textbf{HumanCLAW-Bench}, which asks
whether current VLMs can carry an abstract, long-horizon task through to
completion.
It instantiates one concrete slice of action intelligence: closed-loop
whole-body decision-making.
The benchmark contains $1{,}218$ egocentric whole-body
\texttt{find-navigate-interact} episodes across $41$ indoor houses: find a
target object, bring the body to it, and sit on it.
Each goal specifies what to achieve but leaves open what the body should do at
every moment, so success depends on a sequence of sub-second action decisions.
The decision maker remains frozen throughout; the benchmark measures how far
internet-learned reasoning extrapolates when each choice becomes physical
motion.

Across nine frontier VLMs, none reliably solves the benchmark: the strongest
model completes the full find--navigate--interact progression in only
$16.8\%$ of episodes.
The results reveal a consistent gap between \textit{seeing} and
\textit{acting through a body}.
Once a target is genuinely rendered in the egocentric view, the strongest
model's reported target-seeing rate comes within $5$ percentage points of the
objectively measured visibility rate.
The failures concentrate in everything after seeing: the agent cannot
reliably determine whether it has arrived, where its body is placed relative
to the scene, or when its motion has collided with the world.
We call this missing capacity \textbf{embodied self-awareness}: the ability to
maintain an online estimate of one's own body state, its relation to the
surrounding scene, and the physical consequences of its recent actions.
Current VLMs can describe the world, but they do not reliably track the body
they now control.
With motor execution factored out by construction, these failures belong to
the decision maker.

In summary, our contributions are threefold:
\begin{itemize}
    \item \textbf{HumanCLAW}, a controlled setting in which an off-the-shelf
    VLM acts freely in the physical world: the decision stays decoupled from
    motor execution, each atomic skill is realized as continuous whole-body
    motion with physical consequences, and every failure reads at the
    decision level---with the decision maker never trained.
    \item \textbf{HumanCLAW-Bench}, $1{,}218$ egocentric whole-body
    \texttt{find-navigate-interact} episodes across $41$ indoor houses with
    staged, progressive metrics.
    \item \textbf{An extensive evaluation and failure analysis} of nine
    frontier VLMs, which locates their bottleneck not in perception but in
    \emph{embodied self-awareness}: knowing where the body is, whether it has
    arrived, and what its motion has just caused.
\end{itemize}

%% file: content/3_method.tex
\begin{figure}
    \centering
    \includegraphics[width=0.9\linewidth]{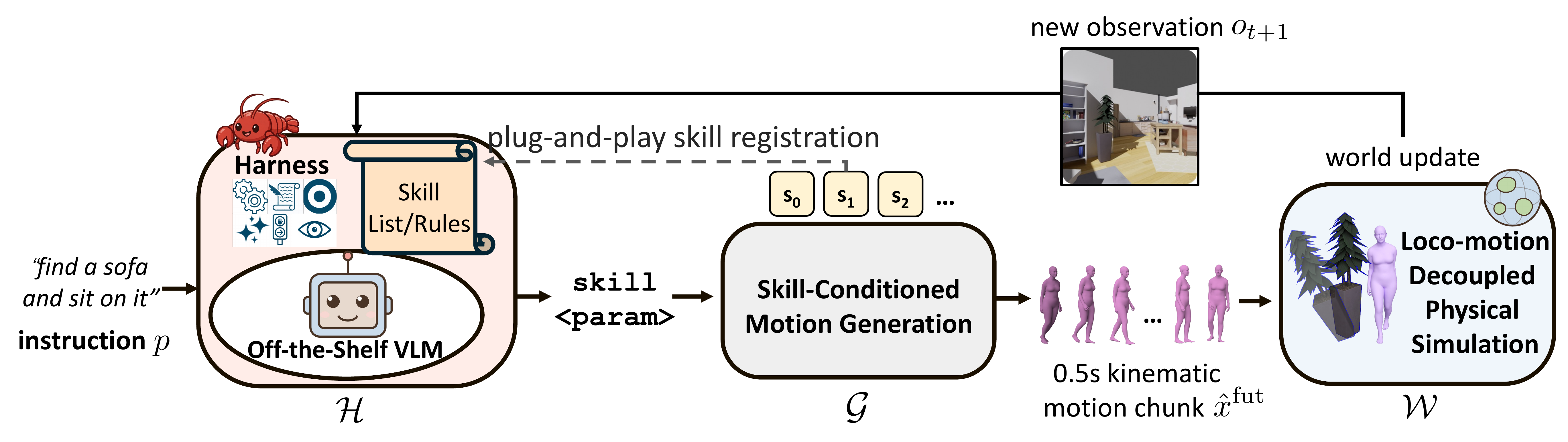}
    \caption{\textbf{HumanCLAW closed-loop framework.} A harnessed off-the-shelf VLM
    maps each egocentric observation to a skill call \texttt{skill<param>}; a
    skill-conditioned generator turns it into a $0.5$\,s full-body motion chunk, which a
    half-physics simulator executes under contact and gravity to return the next
    observation. New skills register plug-and-play, without retraining the VLM.}
    \label{fig:pipeline}
\end{figure}
\section{HumanCLAW}
\label{sec:method}

\subsection{Problem Setting}

CLAW-like agents make language goals actionable through symbolic interfaces.
HumanCLAW studies the same agentic loop when the interface is a full body in
physical space. The agent still observes, decides, acts, and receives feedback,
but an action no longer ends as a command: it is realized as body motion and
constrained by the physical world.

Concretely, HumanCLAW pursues a long-horizon task specified by a single high-level
instruction $p$ (e.g., \textit{``find a sofa and sit on it''}) that fixes the goal
but leaves every execution detail open. As shown in Figure~\ref{fig:pipeline}, it
closes this loop at a sub-second rate. At each timestep $t$, the agent takes a
2D egocentric RGB observation $o_t$ and a compact textual history $h_t$ of past
decisions and feedback, and a skill-decision module built around a harnessed VLM
maps them to a single parameterized skill call
\begin{equation}
\langle s_t, c_t \rangle = H(p, o_t, h_t),
\end{equation}
where $s_t$ is an atomic skill from a fixed skill set and $c_t$ its continuous
parameter. A decoupled motion generator $G$ then realizes the call as a sub-second,
zero-shot-reliable future motion chunk
\begin{equation}
\hat{x}^{\mathrm{fut}}_t = G(x_t, s_t, c_t),
\end{equation}
conditioned on the current body state $x_t$. Finally, a locomotion-decoupled physical
simulator $W$ executes this chunk, updates the physical state of the world, and
returns the next observation $o_{t+1} = W(x_t, \hat{x}^{\mathrm{fut}}_t)$.

Observe--think--act thus forms a complete closed loop: rather than committing to a
fixed plan, the agent dynamically decides what to do next and continually adjusts to
the spatial relation between its body and the scene that its own actions reshape.

\subsection{Egocentric Action Decision}
                                                                                                      
\begin{figure}[h!]                                             
  \centering                                                  
  \includegraphics[width=0.9\linewidth]{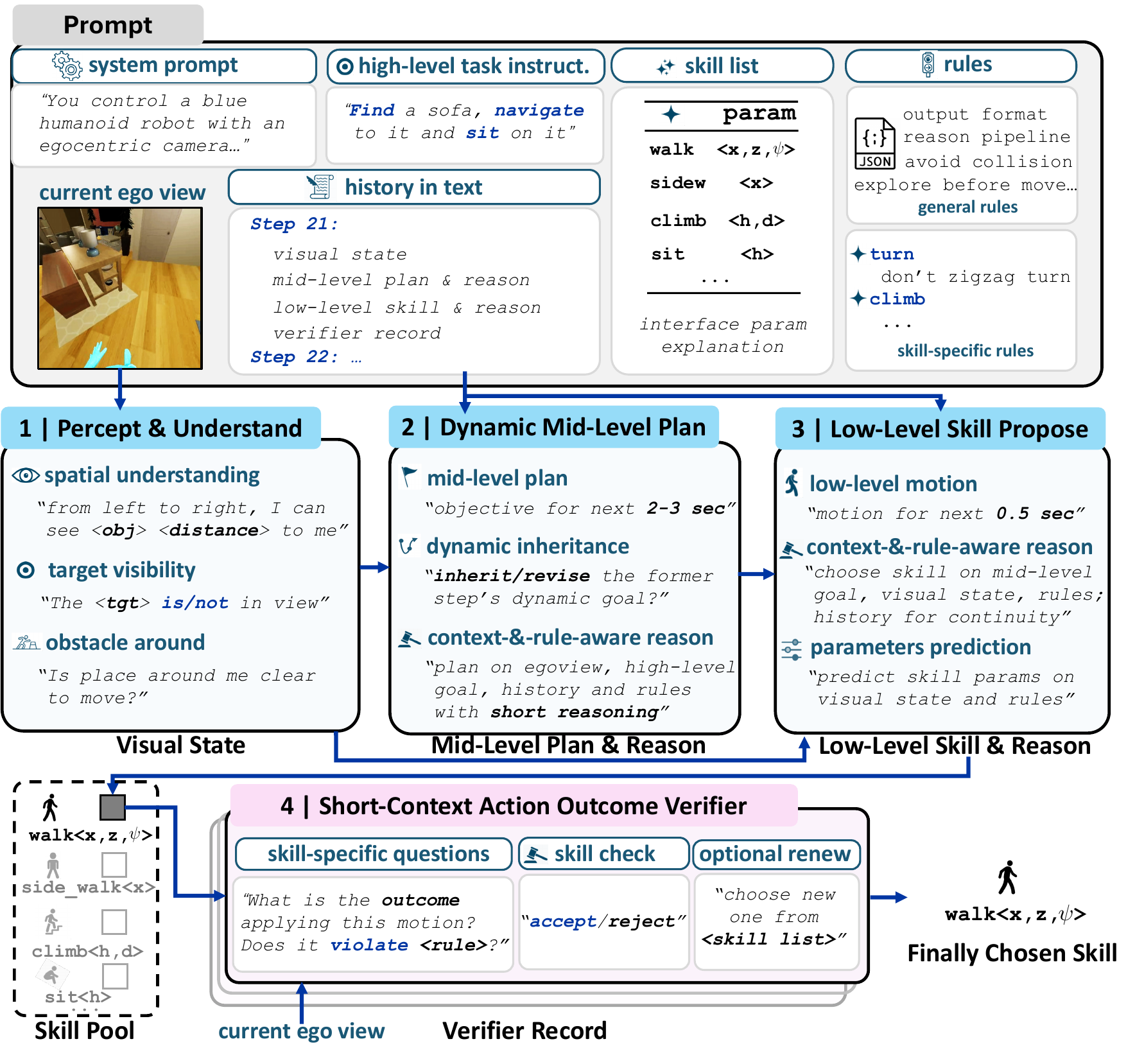}          
  \caption{HumanCLAW harness. We guide an open-ended VLM from egocentric  
  perception to executable humanoid control through a contextual         
  high-to-mid-to-low reasoning scaffold, followed by a skill-specific verifier
  that corrects spatially unsafe or premature skill proposals.}
  \label{fig:humanclaw_harness}                
\end{figure}                       
                               
As shown in Fig.~\ref{fig:humanclaw_harness}, HumanCLAW employs a contextual reasoning harness that enables an open-ended VLM to infer the low-level motion skill a humanoid should execute over the next $0.5$s from egocentric visual observations. Rather than prompting the VLM to predict motion directly, the harness progressively structures its decision-making through guided reasoning. It grounds the model in the agent’s embodiment and task context, elicits an explicit ego-centric spatial state, guides reasoning from high-level intent to mid-level decisions and low-level skills, and verifies the proposed skill before execution.

  \textbf{Contextual Prompting.}
  At each step, the VLM receives a prompt that specifies the agent identity, the
  high-level instruction, the available motion skills, and the rules for using
  each skill. For example, the model is told that it controls a blue humanoid
  robot with an egocentric camera, and that its high-level goal is to find a target
  object and navigate to or interact with it, such as finding a toilet and
  sitting on it. The prompt also includes a compact \textit{textual history} of recent
  steps, including previous visual states, reasoning traces, selected actions.
  This history is a key cue for action coherence: it informs
  the VLM what it has been trying to do, how the environment responded, 
  rather than treating each egocentric image as an independent captioning problem.

  \textbf{Ego-Perception and Spatial Understanding.}
  Before choosing a motion, the VLM is required to output a formatted visual
  state from the current egocentric observation. This state describes visible
  objects from left to right, their approximate distance to the body, whether the
  target object is visible, and whether the immediate surrounding area contains
  obstacles or relevant affordances such as stairs, open space, or a sitting
  surface. This step externalizes the model's spatial interpretation, making the
  subsequent decision depend on an explicit ego-view understanding rather than an
  implicit image-to-action guess.

  \textbf{High-to-Mid-to-Low Reasoning.}
  The core of the harness is a progressive high-to-mid-to-low reasoning scaffold.
  The VLM first interprets the long-horizon instruction under the current visual
  state, e.g., whether the agent should search for the target, approach it, align
  with it, or complete an interaction. It then derives a mid-level objective that
  may remain valid for several seconds, such as moving toward a visible object,
  entering an open area, aligning with stairs, or turning before sitting. Finally,
  the VLM converts this mid-level objective into the low-level motion skill to
  execute at the current step.
  This differs from treating the mid-level objective as an open-loop plan. At
  every step, the VLM  explicitly decide whether to inherit the previous
  mid-level objective or revise it according to the new egocentric observation and
  feedback. The harness therefore preserves the stability of hierarchical
  planning while keeping the policy closed-loop and reactive.

  \textbf{Low-Level Motion Skill Choice.}
After spatial understanding and mid-level reasoning, the VLM selects a motion skill from a predefined skill pool and predicts its associated parameters. 
Specifically, it  infers both the skill category (e.g., \texttt{turn} or \texttt{sit}) and its continuous parameters (e.g., \texttt{<\textit{degree}>} or \texttt{<\textit{height}>}). 
The resulting action is emitted in a structured JSON format, providing the motion generator with a deterministic interface for instantiating the selected skill.

  \textbf{Skill-Specific Spatial Response Verifier.}
  Before motion generation, HumanCLAW applies a short-context verifier specialized
  to the proposed skill. The verifier checks whether executing the proposed
  motion in the current spatial state would cause an unintended consequence: for
  example, walking into an obstacle, stopping before the target is reached,
  climbing without a visible stair, or sitting from an invalid pose. This step is
  motivated by an observation that VLM spatial reasoning degrade as rollout
  context grows longer; the planner may hallucinate progress, such as claiming it
  has reached an object that remains far away. The verifier uses compact,
  skill-specific prompts to decide whether to accept the proposal or replace it
  with a corrected skill from the same action pool; the per-skill checklist is
  given in \Cref{tab:app_verifier}.

\begin{figure}[t]                                                                                            
  \centering                                                                                               
  \includegraphics[width=0.92\linewidth]{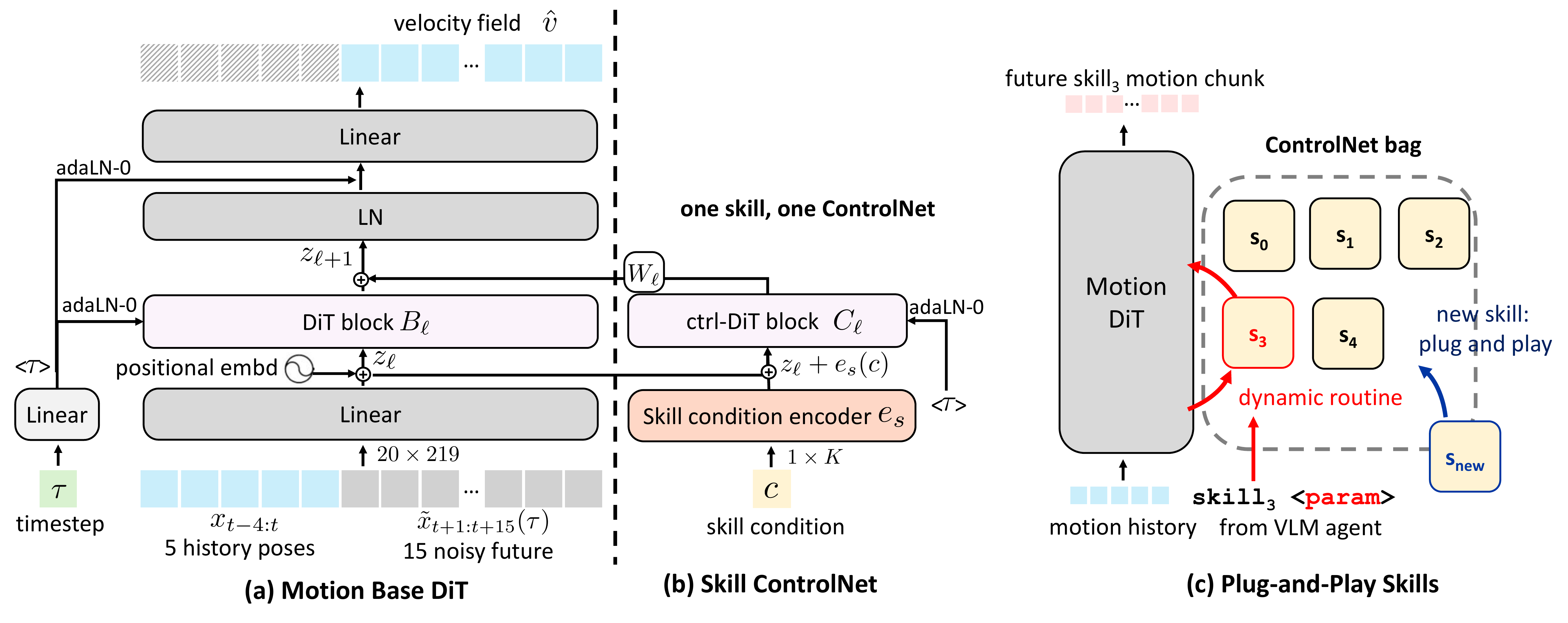}          
  \caption{\textbf{Skill-conditioned motion generator.} \textbf{(a)} A skill-agnostic
  Motion Base DiT predicts the velocity field of $15$ future frames from $5$ history poses
  via flow matching. \textbf{(b)} Each skill is a ControlNet adapter injected through a
  zero-initialized $W_\ell$, leaving the base frozen. \textbf{(c)} Adapters form a
  plug-and-play bag: the VLM-selected skill routes to its own ControlNet, and new skills
  add one more without touching the rest.}
  \label{fig:motion}                
\end{figure} 

\subsection{Skill-Conditioned Motion}

Real human actions are fundamentally different from hard-coded actions in a
symbolic agent. A code agent can turn left and immediately turn right, or switch
from walking forward to walking backward without any transition. Human motion
cannot: a person needs to decelerate, shift weight, recover balance, and produce
short transient motions between intentions. 
These unavoidable ``imperfections'' are
part of embodied action intelligence. 
Action decision should therefore be executed and evaluated under the physical consequences of natural human motion, not under an idealized instantaneous controller.

Meanwhile, however, the decoupled design
requires a reliable zero-shot interface between the VLM skill selector and the motion
actor. If the planner issues \texttt{walk forward}, the generated motion should
correspond to a commonsense human walk: it should not moonwalk, stagger, or
produce unrelated upper-body motion. Otherwise, failures can no longer be
attributed to the planner's action intelligence, because the motion interface
itself is misaligned with the semantic meaning of the skill.

\textbf{Atomic Skills with Deterministic Text Interface.}
We address this tension by defining a finite set of atomic, semantically
unambiguous motion skills. The VLM agent is responsible for composing
these skills over time, while the motion generator is responsible only for
realizing the selected primitive. For example, sitting on a sofa is \textit{not} exposed
as a single command such as \texttt{sit(sofa at (x,z))}. Instead, it is the VLM agent that
decides whether to walk toward the sofa, turn, step backward, or sit based on the
current egocentric state. This keeps long-horizon composition in the action
intelligence module while keeping each motion primitive locally reliable.

  \begin{tcolorbox}[          
      colback=gray!8,
      colframe=gray!25,                                                                                                                                                                                                        
      boxrule=0.5pt,                                                                                                                                                                                                           
      arc=2pt,                                                                                                                                                                                                                 
      left=6pt,    
      right=6pt,
      top=4pt,
      bottom=4pt
  ]
  \[
  \begin{aligned}
  &\texttt{walk}(x, z, \psi), \quad
  \texttt{side\_step}(x), \quad
  \texttt{step\_back}(z), \quad
  \texttt{turn\_in\_place}(\theta), \\
  &\texttt{climb\_upstairs}(h, d), \quad
  \texttt{walk\_downstairs}(h, d), \quad
  \texttt{sit\_in\_place}(h), \quad
  \texttt{stop}.
  \end{aligned}
  \]
  \end{tcolorbox}
Here $(x,z,\psi)$ specifies local walking displacement and heading change,
$\theta$ is the signed turning angle, $(h,d)$ denotes stair height and depth,
and $h$ denotes the target sitting height. Each skill has a deterministic text
and JSON interface for the VLM, but its execution is generated as continuous
human motion rather than retrieved as a fixed animation.

This design gives HumanCLAW both sides of the decoupling. From the VLM's
perspective, the action space is stable, interpretable, and commonsense-aligned:
a skill name has a predictable local outcome. From the simulator's perspective,
the executed trajectory remains a realistic full-body motion with natural
transitions, contact-relevant body pose, and variation across contexts. The
motion generator below implements this interface by using a general motion DiT
as a human motion prior and attaching skill-specific ControlNet adapters for
parameterized control.

 This interface requires motion generation to be both realistic at the trajectory
  level and reliable at the skill level: motions should contain natural
  transitions and variation, while each skill name consistently produces its
  intended local outcome.

\textbf{Motion Base DiT.}
Unlike most motion generation models that synthesize a complete sequence, our
backbone follows \cite{zhang2025primal} and is trained for receding-horizon continuation. At step $t$, it observes
the last five clean body states and generates the next fifteen frames at
$30$ fps, corresponding to the $0.5$s motion chunk executed by the simulator. The
base model is intentionally skill-agnostic: it inherits the current motion
dynamics, but not the skill name or task intent. It therefore serves as a
general prior over locally natural human motion, while semantic control is
added later through skill ControlNet.

A key design choice is how the past motion state conditions generation. Instead
of compressing history into a single condition vector~\citep{zhang2025primal}, we concatenate it
directly with the noisy future as motion tokens:
\begin{equation}
u_\tau =
\left[x_{t-4:t}, \tilde{x}_{t+1:t+15}(\tau)\right]
\in \mathbb{R}^{20 \times d}, \quad d=219.
\end{equation}
Each frame state is linearly projected to a $38$M-parameter, $10$-layer DiT with $8$-head self-attention processes the full
  $20$-frame sequence. The flow time $\tau$ is
embedded by a sinusoidal MLP and injected through adaLN-Zero in every DiT block;
the base model uses no history encoder and no semantic condition. The network
predicts a velocity field for all tokens, and we keep only the future-token
outputs:
\begin{equation}
\hat{v}_\theta =
G_\theta\!\left(
\left[x_{t-4:t}, \tilde{x}_{t+1:t+15}(\tau)\right], \tau
\right)_{t+1:t+15}.
\end{equation}

\begin{figure}
    \centering
    \includegraphics[width=0.98\linewidth]{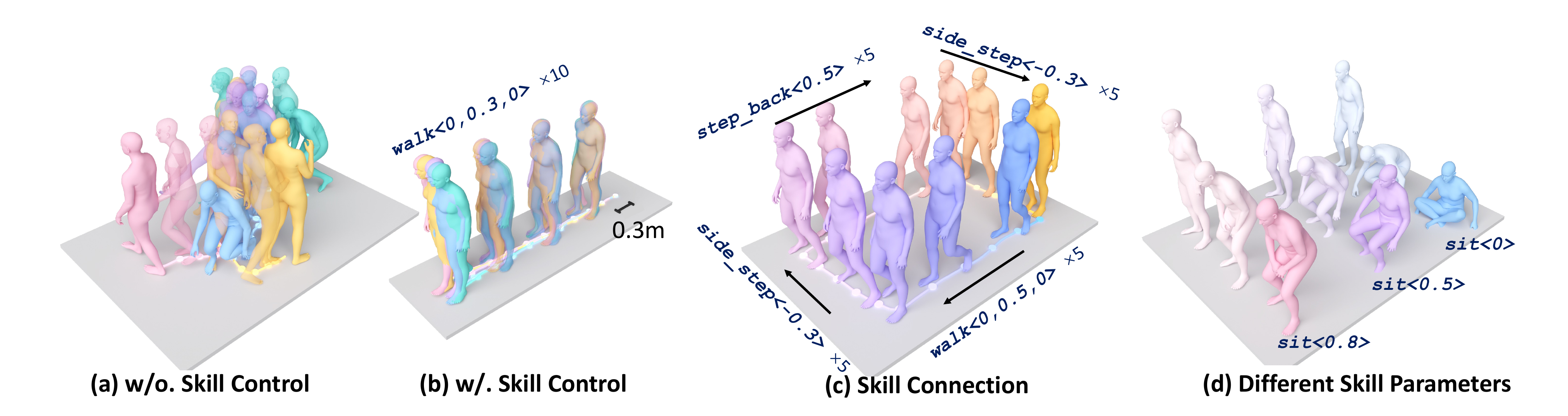}
    \caption{\textbf{Skill controllability.} \textbf{(a)} Without skill control the prior
    collapses into colliding poses; \textbf{(b)} with control, \texttt{walk} tracks the
    requested $0.3$\,m stride. \textbf{(c)} Skills compose with natural transitions
    (\texttt{walk}/\texttt{side\_step}/\texttt{step\_back} into a loop); \textbf{(d)}
    \texttt{sit} at heights $0$/$0.5$/$0.8$ realizes the commanded parameter.}
    \label{fig:motion_result}
\end{figure}

This history-as-token conditioning makes kinematic continuity explicit: future
  tokens can directly attend to the exact recent pose and velocity states, rather
  than relying on a compressed history representation that may lose boundary
  information.

We train the backbone on AMASS motions preprocessed into $20$-frame chunks at
$30$Hz. Before training, each segment is calibrated into a chunk-local frame
whose origin and heading are defined by the initial body state. This removes
global translation and yaw, so the base DiT models local human dynamics rather
than absolute world motion. The first five frames are clean history, and the
last fifteen frames form the clean future chunk
$x^{\mathrm{fut}} = x_{t+1:t+15}$. Flow matching samples Gaussian noise
$\epsilon \sim \mathcal{N}(0,I)$ and a time $\tau \in [0,1]$, constructs
\begin{equation}
\tilde{x}_{t+1:t+15}(\tau)
= \left(1-(1-\sigma_{\min})\tau\right)\epsilon
+ \tau x^{\mathrm{fut}},
\quad
v^\star = x^{\mathrm{fut}} - (1-\sigma_{\min})\epsilon,
\end{equation}
and minimizes $\|\hat{v}_\theta - v^\star\|_2^2$ on the future frames. At
inference, we start from noise and solve the learned ODE with a $30$-step
midpoint solver. The resulting base DiT provides a skill-agnostic motion prior:
it produces motion that is kinematically continuous with the past, while the
next stage aligns this prior with the text-level skill name and its continuous
parameters.

\textbf{Plug-and-Play Skills as ControlNet.}
Once the base DiT provides a kinematically continuous motion prior, the next
problem is to align this prior with the zero-shot skill interface exposed to the
VLM. As shown in Figure~\ref{fig:motion}, we do this with one ControlNet-style adapter per atomic skill, while
keeping the base DiT frozen. Thus, adding a new skill only involves training a new ControlNet and does not change any previously trained skill. At rollout time,
the VLM-selected skill name simply chooses which adapter to attach, and the
skill parameters provide the adapter condition.

For a skill $s$ with continuous parameter $c_t$, we copy the DiT blocks of the
base model into a trainable control branch and add a skill-condition encoder
$e_s(c_t)$. The encoder is chosen according to the parameter type: scalar
conditions such as side-step distance or sitting height use an MLP, while
spatial conditions such as walking displacement and yaw use Fourier features
before projection to the DiT hidden dimension. At layer $\ell$, the frozen base
branch and trainable control branch are combined as
\begin{equation}
z_{\ell+1}
= B_\ell(z_\ell, \tau)
+ W_\ell C_\ell(z_\ell + e_s(c_t), \tau),
\end{equation}
where $B_\ell$ is the frozen base block, $C_\ell$ is the trainable control
block, $e_s$ is the skill-condition encoder, and $W_\ell$ is initialized to zero. 
The final projection layer remains the one
from the base DiT; the adapter only modifies the hidden token stream.

The skill parameterization is deliberately tied to physical outcomes that can be
read from the motion chunk itself. For walking, the condition is the
chunk-local final pelvis displacement and yaw change. For side stepping and
stepping backward, it is the corresponding lateral or backward displacement.
For turning, it is the final body yaw. For stairs and sitting, the parameters
are heuristic geometric quantities such as step height/depth or target sitting
height. Therefore, training does not require manually labeled action
parameters: after we curate or filter AMASS clips for a skill, the continuous
condition is extracted from the calibrated chunk geometry. This does not mean
that the training data is used indiscriminately. We manually review and
aggressively filter the clips for each ControlNet to keep the skill distribution
single-purpose and commonsense-aligned. Clips with abnormal style or mixed
intent, such as unstable walking, walking while performing unrelated upper-body
actions, or stair motions with excessive body twisting, are removed. This
curation is important for zero-shot alignment: when the VLM calls
\texttt{walk}, \texttt{turn}, or \texttt{climb\_upstairs}, the corresponding
adapter should realize the ordinary meaning of that skill rather than an
idiosyncratic AMASS motion.

Each adapter is trained with the same flow-matching objective as the base DiT.
The base network is frozen, and only the condition encoder, copied control
blocks, and zero-initialized residual projections are optimized. Given a clean
history, a noisy future, and the extracted skill condition, the adapter predicts
the future-frame velocity field and is supervised with
$\|\hat{v}_{\theta,s}-v^\star\|_2^2$. We train skill adapters with AdamW using
learning rate $3\times10^{-4}$, batch size $2048$, and $5\times10^5$ to
$1.5\times10^6$ optimization steps depending on the skill. We do not use
classifier guidance or classifier-free guidance; skill control is provided
directly by the selected adapter and its deterministic parameter interface.

Figure~\ref{fig:motion_result} shows these properties qualitatively. On its own the
base DiT produces diverse but unconstrained trajectories \textbf{(a)}; attaching the
skill ControlNet converges them to the zero-shot-reliable behavior the skill name
denotes \textbf{(b)}. Because all skills share the same motion backbone, transitions
between consecutive skills stay smooth, with inertia consistent with human
dynamics \textbf{(c)}. Moreover, as the continuous condition is read directly from
motion geometry, this unsupervised parameterization affords fine-grained control over
the skill outcome, e.g., sitting to different target heights \textbf{(d)}.

This design makes the skill set extensible. Adding a new primitive only
requires defining its text/JSON schema, collecting or filtering motion clips
with the relevant outcome parameter, and training a new adapter on top of the
frozen base. Conversely, improving one skill adapter does not affect the VLM
prompt format, the simulator, the base prior, or the other skills. 

\begin{figure}
    \centering
    \includegraphics[width=0.95\linewidth]{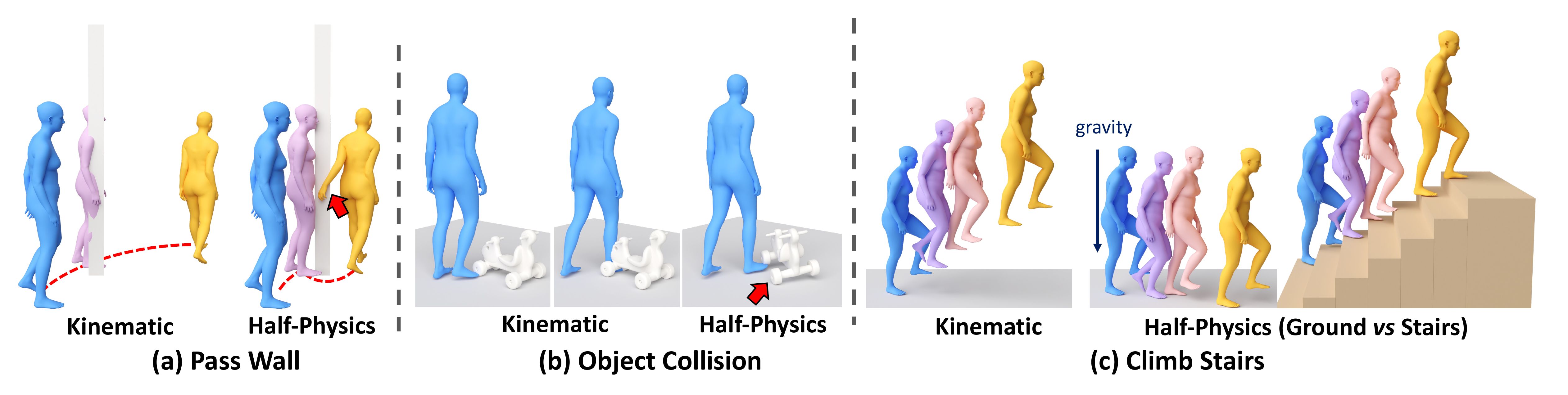}
    \caption{\textbf{Half-physics versus pure kinematics.} Pure kinematic playback walks
    \emph{through} a wall and climbs stairs in mid-air; half-physics subjects the same
    motion to contact and gravity---the wall blocks the body and the stairs support
    it---while still factoring out balance and motor-tracking failures.}
    \label{fig:halfphysics}
\end{figure}

\subsection{Locomotion-Decoupled Physical Simulation}

Another challenge in studying embodied action intelligence is how to decouple
action reasoning from low-level locomotion motor control. In a fully dynamic humanoid
simulation, long-horizon task failure can arise from
the motor system. For example, in cluttered environments or on stairs, a humanoid may lose balance and fall, prematurely terminating the episode and preventing a meaningful evaluation of its action intelligence.

At the same time, action intelligence must be evaluated through physical outcomes. Existing code agents often behave more like ghosts: they follow predefined kinematic animation trajectories, bypassing the physical effects afforded by a humanoid's full action space. For a humanoid, however, whether an action is appropriate depends on its effect in the world:
the humanoid may safely pass through a narrow space, knock over a movable
object, be blocked by a wall, or stop before reaching the target. A purely
kinematic simulator removes these consequences, while a fully dynamic torque
controller entangles them with balance and motor-skill failures. HumanCLAW
therefore adopts an intermediate \textbf{\emph{half-physics}}~\citep{siyao2025half} simulation mechanism that
preserves physical interaction with the scene while abstracting away low-level
motor control.

In HumanCLAW simulation, the world follows rigid-body physics, including
collision, friction, gravity, articulated objects, and movable objects. The human
body, however, is not actuated by simulated joint torques. Instead, it is driven
by equivalent kinematic velocities derived from the generated motion sequence.
This means the agent does not fail because of balance loss or imperfect motor
tracking, e.g., while stepping on stairs, but still physically interacts with
the environment: a wall can block the body, contacts can push movable objects,
and unsupported vertical motion remains constrained by gravity and contact.

Concretely, we build our environment on AI Habitat~\citep{savva2019habitat} with the Bullet~\citep{coumans2015bullet} physics
engine. Given a generated $0.5$s pose sequence from Sec.~\ref{sec:method}, we
convert consecutive poses into equivalent root and joint velocities. For two
adjacent poses $q_t$ and $q_{t+1}$ with simulation step size $\Delta t$, the
linear joint velocity is computed as
\[
    \dot{q}_t = \frac{q_{t+1} - q_t}{\Delta t},
\]
with rotational components computed from relative rotations in the corresponding
tangent space. These equivalent velocities are applied as the only actuation
signal for the humanoid. Following prior passive-body simulation practice~\citep{siyao2025half}, we
run the simulator at $120$ Hz and use passive joint stiffness $\lambda=1.0$ to
preserve compliant contact response.

This half-physics design lets HumanCLAW evaluate action intelligence under
physical consequences without making the benchmark dominated by locomotion
controller failures. The generated motion specifies the intended body movement,
while the simulator determines how that movement interacts with the physical
world. As a result, failures such as walking into a wall, disturbing objects, or
stopping too early are attributable to action-level decisions, while failures
from low-level balance control are largely factored out.

%% file: content/4_hcb.tex
\section{HumanCLAW-Bench}

\subsection{Task Design and 3D Scene Setup}

\begin{figure}[t]
  \centering
  \includegraphics[width=\linewidth]{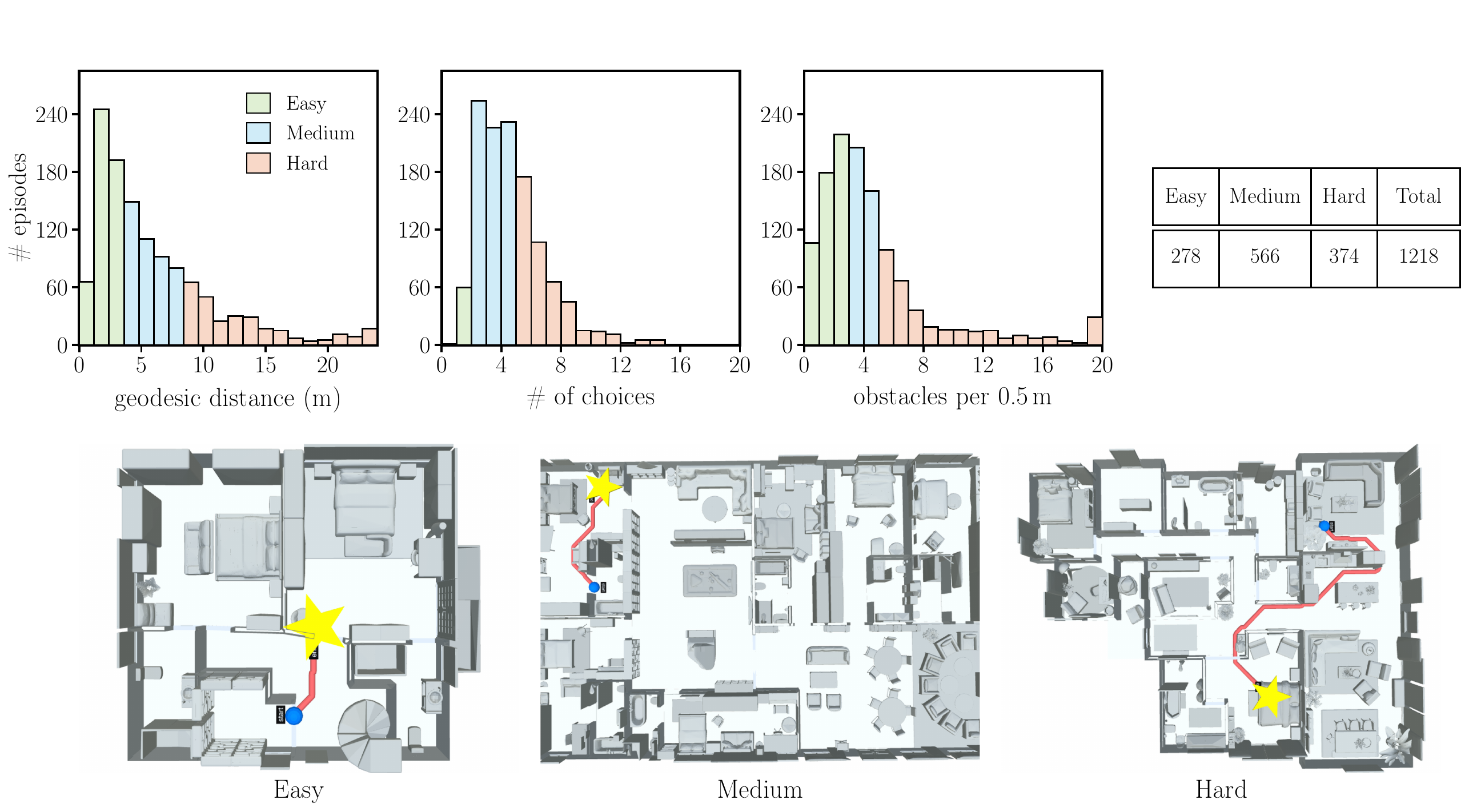}
  \caption{\textbf{Difficulty distributions over the $1{,}218$ benchmark
  episodes.} Each episode is scored along three geometric dimensions---distance
  (geodesic length to the nearest target), choice (turns plus rooms traversed),
  and obstacle (average obstacles within $1\,$m of the route)---and split into
  easy/medium/hard by fixed thresholds; bar color encodes the tier
  (green/blue/pink) and the inset table lists the overall
  counts. Distance is right-skewed, while choice and obstacle spread smoothly,
  giving well-separated difficulty tiers.}
  \label{fig:difficulty-hist}
\end{figure}

We design a \texttt{find-navigate-interact} task to probe \emph{long-horizon action
intelligence}---controlling a body, acting decisively at the $0.5$\,s timescale, and
staying aware of how one's own actions reshape the scene. At the start of each episode
the agent receives a single instruction, \textit{``find a \texttt{<obj>}, navigate to
it with zero distance, and finally sit on it.''} The task is \emph{progressive}: each
stage can be attempted only after its prerequisite is met, so the agent is left to
explore and decide its actions autonomously. We currently instantiate the interaction as
\texttt{sit} (e.g., sitting on a couch or a bed).

We build our 3D scenes on \textbf{HSSD}~\citep{khanna2024hssd}, a large and diverse set of
human-designed indoor houses that span multiple floor levels and a wide variety
of furniture and small objects. Our benchmark uses the $41$ validation houses,
which provide $1{,}218$ episodes over six target categories (\texttt{chair},
\texttt{bed}, \texttt{couch}, \texttt{potted\_plant}, \texttt{toilet},
\texttt{tv}); each episode specifies an agent starting pose and a target object
category, and is solved by reaching any instance of that category.
The six categories split into two subsets: \texttt{bed}, \texttt{couch}, and
\texttt{toilet} form the $597$-episode \emph{sit} subset, whose episodes carry
the full find--navigate--interact progression, while \texttt{chair},
\texttt{potted\_plant}, and \texttt{tv} form the $621$-episode
\emph{navigation} subset, which terminates at navigation.
Find and navigation metrics are therefore computed over all $1{,}218$
episodes, and interaction metrics over the $597$ sit episodes.
To expose the
agent to physical consequence, we further convert plausibly movable objects
(e.g.\ table-top items, small furniture) from static to \emph{dynamic}, so that
contact displaces them; large fixtures (beds, sofas, toilets, walls) remain
static.

\paragraph{Difficulty Analysis.}
To enable difficulty-stratified evaluation, we score every episode along three
interpretable geometric dimensions of the route to the target: \textbf{distance}, the
dataset geodesic distance to the nearest same-category goal; \textbf{choice}, the
navigational branchiness, given by the number of $>\!30^{\circ}$ turns on the
Douglas--Peucker-simplified path plus the number of rooms it traverses; and
\textbf{obstacle}, the clutter, given by the average number of obstacles within
$1\,$m of the route. Since HSSD provides only a scalar geodesic distance and no path
geometry, we first reconstruct the shortest navigable route by running A$^\star$ on a
per-floor $5\,$cm 2D occupancy grid of walls and static furniture; its length matches
the ground-truth geodesic to within a couple percent (median ratio $0.99$), confirming
the routes are faithful. Each dimension is split into \emph{easy}/\emph{medium}/\emph{hard}
by fixed physical thresholds (distance $\le\!3.5$/$\le\!8\,$m; choice $\le\!2$/$\le\!5$;
obstacle $\le\!2.5$/$\le\!5$) that spread the episodes across tiers (roughly
$23$/$46$/$31\%$) rather than forcing equal thirds, and the \textbf{overall} difficulty
averages the three levels. Fig.~\ref{fig:difficulty-hist} reports the resulting
distributions with one worked example per tier, letting us analyze performance and each
geometric factor separately across easy, medium, and hard episodes.

\subsection{Evaluation Metrics.}
We evaluate VLMs along four dimensions: (1) high-level task success, (2) low-level action quality, (3) body awareness and physical disturbance, and (4) computational cost.

\textit{Progressive Success Rate.}
The \texttt{find-nav-interact} benchmark is a progressive task: the agent can proceed to the next subtask only after completing the previous one. We therefore define staged success as both objectively achieved and subjectively acknowledged. That is, the agent needs to satisfy the geometric success criterion and also recognize that the subtask has been completed.
The objective criteria are defined as follows. \textbf{FindSR} is achieved when the target semantic ID appears in the ego-rendered image and occupies at least $100$ pixels in the $512\times512$ view. \textbf{NavSR} is achieved when the minimum distance between the human agent and the target object's axis-aligned bounding box (AABB) is less than 20 cm. \textbf{InteractSR} is achieved when the agent sits down and the pelvis link, corresponding to the hip region, makes contact with the target. Subjective completion is determined from the agent's step-wise visual state and its active decision to stop after completing navigation or interaction; the headline metrics in our tables require both the objective criterion and this subjective completion, while \textsc{Geo}-prefixed variants report the objective criterion alone.

\textit{Action Quality.}
A VLM agent should compose motion skills into smooth and coherent behavior. Poor skill orchestration can lead to unstable motions, such as frequent alternating left and right turns. We measure action smoothness using motion jerk, i.e., the third-order derivative of the agent's kinematic motion.

\textit{Body Awareness and Physical Disturbance.}
Body awareness is a key component of embodied action intelligence. Lack of body awareness can cause unexpected interactions with the environment, such as scraping against a wall while passing through a corridor or knocking over a vase. We therefore report the average number of steps per episode in which a collision occurs, denoted as \textbf{Colls.}.
We also measure environmental disturbance by counting the average number of movable objects that are directly or indirectly displaced by the human agent within an episode, denoted as \textbf{\#Dtb}, and the average displacement distance of each disturbed object, denoted as \textbf{dDtb}.

\textit{Cost.}
Finally, we report the average number of tokens consumed by the agent per episode, including both input and output tokens, as a measure of computational cost.

%% file: content/5_experiment.tex
\section{Experimental Result}
\label{sec:experiment}

\subsection{Zero-Shot Motion Fidelity}

\paragraph{Low-level skills are zero-shot reliable.}
Our controllable motion prior---a shared base diffusion transformer with a per-skill ControlNet, trained
once on AMASS/BABEL and queried \emph{zero-shot} at deployment---faithfully realizes the commanded
direction and magnitude of every skill without any per-task or per-scene tuning. In
Table~\ref{tab:skill_fidelity}, commanding a skill with a target magnitude yields an achievement ratio
close to unity with small variance: walking tracks the commanded step distance at $0.97$, turning the
commanded angle at $0.99$, side-stepping at $1.00$, stepping back at $0.99$, and sitting reaches the
commanded pelvis height at $0.98$, while stair climbing realizes $0.74$--$0.79$ of the commanded riser
as a stable, calibratable gain. The per-command standard deviation is tiny ($\le\!0.12$), so the
controller executes the same command the same way every time---the near-constant ratios reflect a fixed,
predictable gain rather than noise. We contrast this with MoMask~\citep{guo2024momask}, a strong text-to-motion model,
prompted with the identical commands rendered in natural language with the magnitude spelled out
(e.g.\ ``\textit{walks forward at $0.8$ meters per second}''). Even with the number in the prompt, MoMask
renders a \emph{generic} motion for each verb instead of the requested magnitude---it walks at a habitual
speed and sits to a habitual chair height regardless of the command---so its fidelity ratios depart from
unity and scatter with an order of magnitude larger variance (standard deviations of $0.3$--$1.3$ versus
$0.02$--$0.12$ for ours). Because our motor layer is this reliable out of the box, the benchmark largely isolates
\emph{high-level embodied reasoning}---where to go, what to perceive, and how to sequence actions---rather
than low-level motor competence, and the failures we analyze are attributable to planning, perception,
and body-awareness rather than to execution of the underlying skills.

\begin{table}
  \centering
  \caption{\textbf{Skill functional fidelity.} Achievement ratio $=$ achieved\,/\,commanded
  (mean$_{\pm\text{std}}$) over rollouts, evaluated under identical conditions for both methods. Our
  controllable prior tracks every commanded magnitude at $\approx\!1$ with small variance, whereas the
  text-to-motion baseline renders a generic motion per verb and ignores the requested magnitude, giving
  ratios that deviate from $1$ with an order of magnitude larger variance. %
  }
  \label{tab:skill_fidelity}
  \resizebox{1\textwidth}{!}{%
  \begin{tabular}{lccccccc}
    \toprule
    Method & \texttt{walk} & \texttt{side\_step} & \texttt{step\_back} & \texttt{turn\_in\_place}
           & \texttt{climb\_upstairs} & \texttt{walk\_downstairs} & \texttt{sit\_in\_place} \\
    \midrule
    Ours   & $0.966_{\pm0.022}$ & $1.002_{\pm0.123}$ & $0.986_{\pm0.056}$ & $0.994_{\pm0.038}$
           & $0.794_{\pm0.019}$ & $0.738_{\pm0.052}$ & $0.977_{\pm0.078}$ \\
    MoMask~\citep{guo2024momask} & $1.921_{\pm1.031}$ & $1.717_{\pm1.234}$ & $2.072_{\pm1.067}$ & $0.608_{\pm0.284}$
           & $0.820_{\pm0.826}$ & $1.540_{\pm0.421}$ & $1.817_{\pm1.279}$ \\
    \bottomrule
  \end{tabular}%
  }
\end{table}

\definecolor{rankone}{RGB}{186,225,255} %
  \definecolor{ranktwo}{RGB}{224,243,255} %
  \begin{table}[t]
    \centering \small \setlength{\tabcolsep}{4pt} \renewcommand{\arraystretch}{1.15}
    \caption{Full validation results on HumanClawBench 
    FindSR = target rendered in the ego view ($\geq$100 semantic px)
    \emph{and} acknowledged in the model's visible-state text. Disturbance /
    action-quality / cost are per-episode-normalized. Collisions = floor-relative non-ground
    collision step fraction. Motion Jerk = root-rigid jerk 
    at  timescale of $\approx 0.27$s, 
    mean over episodes; lower = more
    coherent motion. \colorbox{rankone}{best} / \colorbox{ranktwo}{2nd} per column.}
    \label{tab:hcb_fullval_nopen}
    \resizebox{\textwidth}{!}{
    \begin{NiceTabular}{l ccc ccc c ccc}
    \toprule
    \multirow{2}{*}{}
    & \multicolumn{3}{c}{High-Level Success Rate}
    & \multicolumn{3}{c}{Body Misaware \& Scene Disturb}
    & \multicolumn{1}{c}{Action Quality}
    & \multicolumn{3}{c}{Cost} \\
    \cmidrule(lr){2-4} \cmidrule(lr){5-7} \cmidrule(lr){8-8} \cmidrule(lr){9-11}
    VLM Backbone
    & FindSR $\uparrow$ & NavSR $\uparrow$ & InteractSR $\uparrow$
    & Coll. $\downarrow$ & \#Dtb/ep\ $\downarrow$ & dDtb(m) $\downarrow$
    & Motion Jerk $\downarrow$
    & avg steps & in tok/step $\downarrow$ & out tok/step $\downarrow$ \\
    \midrule
    GPT-5.5            & 55.1\% & 13.9\% & 3.4\% & 43.4\% & 1.53 & \cellcolor{ranktwo}1.56 & \cellcolor{rankone}4.2 & 82.4 & \cellcolor{rankone}4360 & 354 \\
    Gemini-3.1      & \cellcolor{rankone}64.9\% & \cellcolor{rankone}42.4\% & \cellcolor{rankone}16.8\% & 39.5\% & 1.50 & \cellcolor{rankone}1.22 & 5.7 & 59.1 & 5890 & \cellcolor{ranktwo}311 \\
    Gemini-2.5      & \cellcolor{ranktwo}58.5\% & 21.6\% & 3.5\% & 37.7\% & \cellcolor{ranktwo}1.35 & 3.03 & 8.7 & 71.3 & 6412 & 401 \\
    Claude-4.8       & 32.6\% & 8.6\% & 1.5\% & 44.2\% & 1.73 & 2.76 & 5.3 & 81.0 & 7047 & 625 \\
    \midrule
    Gemma-4-31B     & 58.1\% & \cellcolor{ranktwo}28.7\% & \cellcolor{ranktwo}11.1\% & 40.1\% & 1.66 & 1.86 & \cellcolor{ranktwo}4.8 & 78.5 & 4632 & 322 \\
    Qwen3.6-27B     & 51.0\% & 20.9\% & 0.2\% & 43.0\% & 1.95 & 2.16 & 6.5 & 81.6 & 4862 & 482 \\
    Qwen3.6-35B-A3B & 44.6\% & 5.8\% & 0.0\% & \cellcolor{ranktwo}34.6\% & 1.45 & 2.38 & 7.4 & 74.0 & 4766 & 477 \\
    Qwen3.5-27B     & 37.8\% & 13.5\% & 0.0\% & \cellcolor{rankone}34.5\% & \cellcolor{rankone}0.93 & 2.38 & 6.8 & 37.6 & 4641 & 473 \\
    InternVL3.5-38B & 46.8\% & 0.8\% & 0.0\% & 51.2\% & 1.90 & 2.56 & 7.2 & 93.0 & \cellcolor{ranktwo}4459 & \cellcolor{rankone}309 \\
    \bottomrule
    \end{NiceTabular}}
  \end{table}

\subsection{Benchmark}
We evaluate nine off-the-shelf VLMs as the frozen decision maker and report the
full validation results in Table~\ref{tab:hcb_fullval_nopen}.

\begin{finding}{1}
No current VLM solves the benchmark.
\end{finding}
The best model, Gemini-3.1, finds and sits on the target in only $16.8\%$ of
episodes, and four of the nine sit on it in at most $0.2\%$ of episodes. Behind this low ceiling is a sharp
stage-by-stage drop that every model shares. FindSR ranges from $32.6\%$ to
$64.9\%$, NavSR from $0.8\%$ to $42.4\%$, and InteractSR from $0\%$ to $16.8\%$, so
most of the task is lost twice, first between finding a target and reaching it, and
again between reaching it and interacting with it.

\paragraph{Seeing a target rarely becomes reaching it.}
We summarize the first drop with a landing rate, the ratio of NavSR to FindSR, which
asks how much of what a model finds it actually reaches. Even Gemini-3.1 lands only
$0.65$ of what it finds, Gemma-4-31B $0.49$, and InternVL3.5-38B $0.02$. Recognizing
a target is common across models. Turning that recognition into a body that stops
next to the target is what separates them, and it is where the benchmark leaves the
most room to improve.

\paragraph{Body awareness and cost should be read together with success.}
Low disturbance can signal inaction rather than control. Qwen3.5-27B records the
lowest object displacement ($0.93$ per episode) and the fewest steps ($37.6$), yet
it barely moves and interacts in $0\%$ of episodes. At the other extreme,
InternVL3.5-38B and Claude-4.8 pair the highest collision fractions ($51.2\%$ and
$44.2\%$) with a collapsed NavSR, the profile of an agent that moves without control.
Spending more tokens does not help either. Claude-4.8 uses the most input and output
tokens per step ($7047$ and $625$) but trails on every success metric, while
Gemini-3.1 leads at a lower cost.

\paragraph{Action quality reflects the coherence of a model's choreography.}
Task success says nothing about \emph{how} an agent moves, so we report a \emph{Motion
Jerk} column---the denoised third time-derivative of the pelvis trajectory at the
decision timescale---measuring the coherence of a backbone's choreography rather than
the fixed gait of the shared generator. By construction it penalizes incoherent
\emph{sequencing}, not the amount of motion: a single skill scores $4$--$6$ (forward
walk $5.2$, turn $4.3$) and a purposeful walk--turn--walk trajectory $5.8$, whereas an
aimlessly sequenced policy reaches $9.7$; fixing the skill content and varying only
order confirms this---five forward-walk chunks then five lateral steps score $4.7$, but
interleaving the same ten chunks yields $7.4$ ($1.6\times$ higher) purely from ordering.
Under this measure (lower is smoother), GPT-5.5 is the most coherent ($4.2$), followed
by Gemma-4-31B ($4.8$), and quality is decoupled from success: Gemini-2.5, despite a
competitive success rate, is the \emph{worst} ($8.7$), oscillating side to side with the
highest turn-reversal rate (about one step in five), while Qwen3.6-35B-A3B is second
worst ($7.4$), adding bouts of in-place spinning.

\paragraph{The strongest open model is close to the frontier.}
Gemma-4-31B ($58.1$/$28.7$/$11.1$) matches or beats every proprietary model except
Gemini-3.1, and it beats GPT-5.5 and Claude-4.8 on all three success rates. Progress
on HumanCLAW-Bench is therefore open to the community rather than gated behind a
single closed model.

\begin{table}[t]
        \centering \small \setlength{\tabcolsep}{4pt} \renewcommand{\arraystretch}{1.15}
        \caption{Ablation on HumanClawBench mini-val 100 episodes.  The remaining columns are over all 100 episodes. Text-history
        length, visual-history frames, and design components are varied from the baseline
        (hist 10, img 1). FindSR uses the offline render-find caliber (target renders $\geq$100 px in
        ego view \emph{and} is claimed with no negation), consistent with the main table. Motion Jerk =
        root-rigid jerk at the decision timescale (stride 8), mean over episodes. VLM backbone is Gemma-4-31B.}
        \label{tab:hcb_ablation_nopen}
        \resizebox{\textwidth}{!}{
        \begin{NiceTabular}{l ccc ccc c ccc}
        \toprule
        \multirow{2}{*}{Setting}
        & \multicolumn{3}{c}{High-Level Success Rate}
        & \multicolumn{3}{c}{Body Misaware \& Scene Disturb}
        & \multicolumn{1}{c}{Action Quality}
        & \multicolumn{3}{c}{Cost} \\
        \cmidrule(lr){2-4} \cmidrule(lr){5-7} \cmidrule(lr){8-8} \cmidrule(lr){9-11}
        & FindSR $\uparrow$ & NavSR $\uparrow$ & InteractSR $\uparrow$
        & Coll.\% $\downarrow$ & \#Dtb\ $\downarrow$ & dDtb(m) $\downarrow$
        & Motion Jerk $\downarrow$
        & avg steps & in tok/step $\downarrow$ & out tok/step $\downarrow$ \\
        \midrule
        \textbf{Baseline (hist 10, img 1)} & 58.0\% & 27.0\% & 18.9\% & 41.4\% & 1.35 & 9.38 & 4.9 & 78.5 & 4676 & 322 \\
        \midrule
        \multicolumn{11}{l}{\textit{Text-history length}} \\
        \quad hist 0    & 65.0\% & 11.0\% & 0.0\%  & 38.6\% & 1.27 & 6.46 & 5.1 & 88.9 & 2582  & 292 \\
        \quad hist 20   & 59.0\% & 27.0\% & 7.5\%  & 42.9\% & 1.18 & 1.56 & 4.8 & 80.3 & 6692  & 326 \\
        \quad hist 50   & 60.0\% & 28.0\% & 7.5\%  & 40.4\% & 1.24 & 6.16 & 4.8 & 77.6 & 10538 & 320 \\
        \quad hist 100  & 56.0\% & 26.0\% & 11.3\% & 37.6\% & 1.08 & 7.23 & 4.7 & 79.7 & 12904 & 318 \\
        \midrule
        \multicolumn{11}{l}{\textit{Visual-history frames}} \\
        \quad img 2     & 57.0\% & 31.0\% & 9.4\%  & 43.0\% & 1.19 & 2.55 & 4.7 & 78.8 & 5234  & 321 \\
        \quad img 5     & 55.0\% & 28.0\% & 7.5\%  & 41.1\% & 1.30 & 8.86 & 5.1 & 82.7 & 6335  & 324 \\
        \quad img 10    & 53.0\% & 13.0\% & 3.8\%  & 34.9\% & 0.95 & 3.19 & 5.2 & 92.3 & 7685  & 303 \\
        \midrule
        \multicolumn{11}{l}{\textit{Design components}} \\
        \quad w/o. verifier           & 51.0\% & 2.0\%  & 0.0\%  & 33.2\% & 0.71 & 10.40 & 4.8 & 47.5 & 4095 & 247 \\
         \quad w/o. mid-level & 62.0\% & 30.0\% & 0.0\% & 35.4\% & 1.22 & 2.08 & 5.0 & 70.3 & 3769 & 264 \\
        \bottomrule
        \end{NiceTabular}}
      \end{table}

\subsection{Ablation Study}

We ablate the harness on a $100$-episode mini-val split, varying the text-history
length, the number of visual-history frames, and the design components around the
baseline (hist $10$, img $1$) in Table~\ref{tab:hcb_ablation_nopen}.

\begin{finding}{2}
Reasoning scaffolds---memory, mid-level objectives, and verification---lift
action intelligence; longer text or visual history does not.
\end{finding}

\paragraph{The verifier is the decisive component.}
Removing the verifier collapses NavSR from $27.0\%$ to $2.0\%$ and InteractSR from
$18.9\%$ to $0.0\%$, while FindSR barely changes ($58.0\%$ to $51.0\%$). Its effect
is on acting, not on seeing. Without the verifier episodes also end far earlier
($47.5$ steps versus $78.5$), which shows an agent that stops or drifts before the
loop closes. The verifier is what turns a stream of skill proposals into a closed
loop that reaches and commits.

\paragraph{Text history is necessary but saturates quickly.}
With no history the agent cannot sustain a plan, and NavSR and InteractSR fall to
$11.0\%$ and $0.0\%$. A short structured memory recovers most of the gap, but
lengthening it does not help. NavSR stays near $26\text{--}28\%$ across histories of
$10$, $20$, $50$, and $100$ steps, and InteractSR is highest at the shortest setting,
while input tokens per step climb from about $4.7$K to $12.9$K. A compact history
captures nearly all of the benefit at a fraction of the cost.

\paragraph{More image frames can hurt.}
Adding a few visual-history frames is roughly neutral, but too many degrade the
agent. At img $10$, NavSR drops from $27.0\%$ to $13.0\%$ and InteractSR from $18.9\%$  to $3.8\%$ while input
tokens rise. The decision model is bottlenecked by reasoning over the current view,
so flooding it with past frames dilutes the signal rather than improving coverage.

\paragraph{Mid-level reasoning carries the long-horizon interaction.}
Removing the mid-level objective barely touches finding or navigating---FindSR
($62.0\%$) and NavSR ($30.0\%$) even edge above the baseline---yet InteractSR
collapses from $18.9\%$ to $0.0\%$. The reason is that sitting is not a single
reactive step but a short compositional routine: the agent has to reach the seat, turn
to align its back to it, and only then sit---a several-second sequence that a purely
step-by-step policy cannot hold together. Without a mid-level objective to bind these
steps, the agent approaches the target but never closes the terminal interaction, and
episodes end earlier ($70.3$ vs $78.5$ steps). Mid-level reasoning is thus what
sustains the medium-horizon commitments that turn navigation into interaction.

\subsection{Error Analysis and Key Findings}

\begin{figure}
    \centering
    \includegraphics[
        width=\linewidth,
        trim=3cm 0cm 3cm 0cm,
        clip
    ]{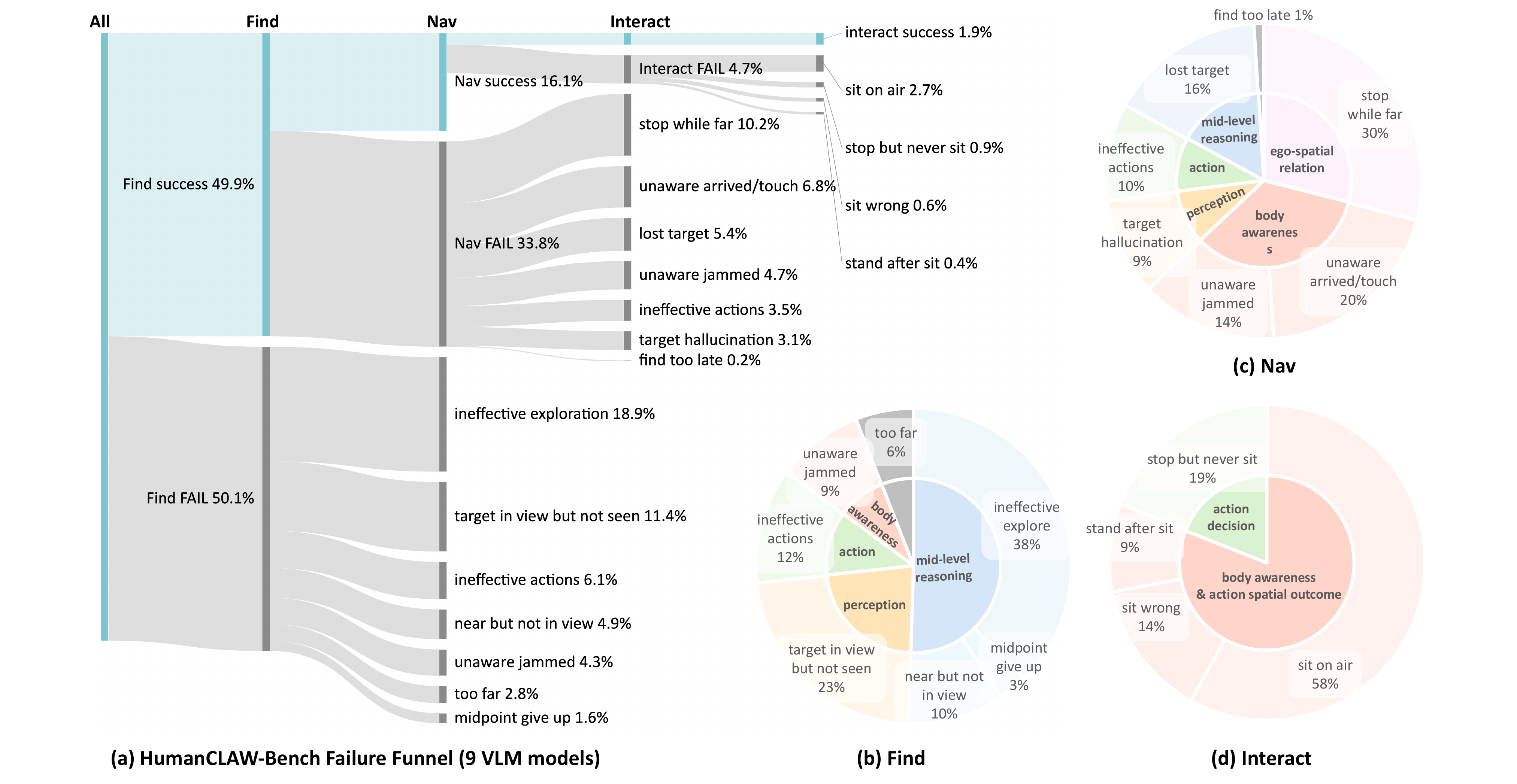}
    \caption{\textbf{Where episodes fail across the nine VLMs.} \textbf{(a)} A
    Find$\to$Nav$\to$Interact funnel over all episodes; \textbf{(b--d)} per-stage
    root-cause breakdowns colored by faculty. Failures concentrate in egocentric body
    awareness and ego-spatial self-localization---stopping while far, unawareness of
    arriving or being jammed, and sitting into thin air---not in perception or motor
    execution.}
    \label{fig:error_analysis}
\end{figure}

Having seen that no current VLM solves the benchmark, we now dissect \emph{where} and
\emph{why} episodes fail, using four complementary views of the same rollouts. A
capability funnel tracks how episodes drain across the find$\to$navigate$\to$interact
stages, and a per-stage breakdown attributes each stage's failures to a single root
cause (Fig.~\ref{fig:error_analysis}); all root-cause labels are assigned
automatically by fixed rules over the logged rollouts (\Cref{app:rootcause}). Table~\ref{tab:hcb_find_nav_variants} reports
success-rate variants that separate \emph{objective} (geometry-only) from
model-\emph{acknowledged} success at each stage, isolating perception from decision;
Table~\ref{tab:hcb_coll_bodygroup} breaks collisions down by body part; and
Fig.~\ref{fig:collision} shows representative body-unawareness rollouts. We organize the
analysis into four findings, one per stage of the embodied loop.

\definecolor{rankone}{RGB}{186,225,255} %
  \definecolor{ranktwo}{RGB}{224,243,255} %
\begin{table}
  \centering \small \setlength{\tabcolsep}{5pt} \renewcommand{\arraystretch}{1.15}
  \caption{Find / Nav / Interact success-rate variants on HumanClawBench full-val. FindSR = render
  ($\geq$100 px in ego view) \emph{and} claimed (no negation); GeoFindSR = render $\geq$100 px only.
  NavSR = active stop \emph{and} body--target AABB distance $\leq$ threshold; GeoNavSR = the AABB distance criterion only.
  InteractSR = a \texttt{sit} action lands the pelvis in \emph{mesh} contact with the target at the
  committed stop; GeoInteractSR = some \texttt{sit} action lands the pelvis in target-mesh contact at
  \emph{any} step (sat on the seat at some point, even if the pose is later lost). \colorbox{rankone}{best} / \colorbox{ranktwo}{2nd} per column.}
  \label{tab:hcb_find_nav_variants}
  \resizebox{\textwidth}{!}{
  \begin{NiceTabular}{l cc ccc cc}
  \toprule
  & \multicolumn{2}{c}{Find} & \multicolumn{3}{c}{Nav} & \multicolumn{2}{c}{Interact} \\
  \cmidrule(lr){2-3} \cmidrule(lr){4-6} \cmidrule(lr){7-8}
  VLM Backbone
  & FindSR $\uparrow$ & GeoFindSR $\uparrow$
  & NavSR@20cm $\uparrow$ & GeoNavSR@20cm $\uparrow$ & NavSR@$1$m $\uparrow$
  & InteractSR $\uparrow$ & GeoInteractSR $\uparrow$ \\
  \midrule
  GPT-5.5     & 55.1\% & 63.2\% & 13.9\% & 27.7\% & 15.5\% & 3.4\% & 9.7\% \\
  Gemini-3.1      & \cellcolor{rankone}64.9\% & \cellcolor{rankone}69.9\% & \cellcolor{rankone}42.4\% & \cellcolor{rankone}44.7\% & \cellcolor{rankone}52.5\% & \cellcolor{rankone}16.8\% & \cellcolor{rankone}21.4\% \\
  Gemini-2.5      & \cellcolor{ranktwo}58.5\% & 66.4\% & 21.6\% & 23.6\% & 31.0\% & 3.5\% & 4.5\% \\
  Claude-4.8      & 32.6\% & 55.7\% & 8.6\% & 13.5\% & 11.4\% & 1.5\% & 2.2\% \\
  \midrule
  Gemma-4-31B     & 58.1\% & \cellcolor{ranktwo}67.7\% & \cellcolor{ranktwo}28.7\% & \cellcolor{ranktwo}35.6\% & \cellcolor{ranktwo}33.4\% & \cellcolor{ranktwo}11.1\% & \cellcolor{ranktwo}17.9\% \\
  Qwen3.6-27B     & 51.0\% & 62.8\% & 20.9\% & 34.8\% & 24.2\% & 0.2\% & 0.3\% \\
  Qwen3.6-35B-A3B & 44.6\% & 59.0\% & 5.8\% & 14.0\% & 11.4\% & 0.0\% & 0.0\% \\
  Qwen3.5-27B     & 37.8\% & 48.9\% & 13.5\% & 17.7\% & 19.2\% & 0.0\% & 0.0\% \\
  InternVL3.5-38B & 46.8\% & 59.1\% & 0.8\% & 12.6\% & 1.6\% & 0.0\% & 0.0\% \\
  \bottomrule
  \end{NiceTabular}}
\end{table}

\begin{finding}{3}
Perception matters but is not the bottleneck.
\end{finding}
Table~\ref{tab:hcb_find_nav_variants} separates two notions of finding: \textbf{GeoFindSR}
counts an episode whenever the target is \emph{objectively} rendered in the ego view
(at least $100$ pixels in the $512\times512$ image), whether or not the model says so,
while \textbf{FindSR} additionally requires the model to acknowledge it. For the
strongest models the two differ by only $5$--$10$ points (Gemini-3.1 $69.9$ vs.\ $64.9$,
GPT-5.5 $63.2$ vs.\ $55.1$, Gemma-4-31B $67.7$ vs.\ $58.1$), so once a target genuinely
enters the ego view it is almost always recognized---recognition itself is largely
solved (Fig.~\ref{fig:error_analysis}a). Where finding breaks is upstream, in bringing
the target into view at all: about half of all episodes never do. Replaying every
failed episode and attributing it to a single root cause (Fig.~\ref{fig:error_analysis}b),
the dominant deficit is \emph{mid-level reasoning}: chiefly \textbf{ineffective
exploration} ($38\%$), where the agent searches the scene incoherently and never
renders the target, together with related search failures---approaching without ever
turning to view ($10\%$) and giving up mid-route ($3\%$). Genuine \textbf{perception}
lapses---the target is rendered but goes unacknowledged---account for a secondary
$23\%$, and the rest are ineffective actions ($12\%$), being unknowingly jammed against
geometry ($9\%$), and abandoning a too-far goal ($6\%$). Finding is therefore gated by \emph{where and how the agent chooses
to explore}, not by whether it can recognize a target once it appears.

\begin{finding}{4}
Egocentric self-localization is the navigation bottleneck.
\end{finding}
Among the 5{,}473 episodes across all nine VLMs in which the agent \emph{actively finds} the
target---the object is rendered with $\geq$100 semantic pixels in the ego view \emph{and} is
acknowledged in the model's visible-state description---3{,}706 (68\%) still fail to navigate to
it (reach within $0.2$\,m and stop). We replay every such trajectory offline and, using the
objective last-frame rendering together with the per-step body displacement and action stream,
attribute each failure to a single root cause (Fig.~\ref{fig:error_analysis}). Nearly two-thirds of the
failures are errors of \emph{egocentric self-spatial awareness} rather than of perception or
low-level control, and they surface as a \emph{termination} decision that fails in both
directions: the agent either keeps navigating after it has already arrived or stops while
still far away, because it cannot tell where its body is relative to the target. The
largest group is \textbf{body awareness} (34\%): the agent is physically
at or against the target yet misjudges its own state---in 20\% it has arrived within $0.2$\,m but
never signals arrival and keeps navigating, and in 14\% it is jammed against an obstacle for a stretch of steps yet keeps
issuing forward commands, unaware it is blocked. A further \textbf{30\%} is an \emph{ego-spatial
distance hallucination}: the agent stops and declares arrival while still more than $0.2$\,m away,
believing a distant target is within reach. The remaining failures are comparatively minor:
\textbf{mid-level reasoning} (17\%; the agent turns away---e.g.\ to avoid an obstacle---loses the
target from view and cannot re-acquire it, 16\%, or finds it too late to physically reach, under
1\%), \textbf{visual perception} (9\%; the target has left the ego view but the model still claims
to see it), and pure \textbf{action/locomotion} (10\%; the body moves freely but keeps turning and
issuing ineffective actions, never converging on the visible target). Discovery \emph{timing} is
negligible (under 1\%): with a distance-aware criterion, only 0.5\% of failures lack enough
remaining steps to cover the distance after first sighting. In short, the humanoid agent's core
weakness is not \emph{whether it can see the target} or \emph{whether it can move}, but
\emph{knowing where it is relative to---and whether it has reached---the target it already sees.}

\begin{figure}[t]
  \centering
  
  \begin{minipage}[c]{0.53\linewidth}
    \centering
    \captionof{table}{Per-body-group collision breakdown on HumanClawBench. \textbf{Coll.\%} = any non-ground body part; the four group columns are per-group step fractions and overlap. \colorbox{rankone}{best} / \colorbox{ranktwo}{2nd} per column.}
    \label{tab:hcb_coll_bodygroup}
    
    \small 
    \setlength{\tabcolsep}{4pt} %
    \renewcommand{\arraystretch}{1.15}
    \resizebox{\textwidth}{!}{
    \begin{NiceTabular}{l ccccc}
    \toprule
    VLM Backbone & Coll.\% $\downarrow$ & Arm/Hand $\downarrow$ & Torso $\downarrow$ & Leg/Foot $\downarrow$ & Head $\downarrow$ \\
    \midrule
    GPT-5.5           & 43.37 & 23.24 & 13.78 & 38.83 & 2.99 \\
    Gemini-3.1      & 39.48 & \cellcolor{ranktwo}20.97 & 11.72 & 33.29 & \cellcolor{rankone}2.62 \\
    Gemini-2.5      & 37.69 & 22.70 & 13.63 & 30.87 & 2.94 \\
    Claude-4.8      & 44.18 & 28.97 & 16.90 & 37.94 & 6.93 \\
    \midrule
    Gemma-4-31B     & 40.09 & 21.21 & \cellcolor{ranktwo}11.38 & 34.72 & \cellcolor{ranktwo}2.66 \\
    Qwen3.6-27B     & 42.97 & 25.48 & 12.70 & 36.86 & 3.54 \\
    Qwen3.6-35B-A3B & \cellcolor{ranktwo}34.58 & 21.20 & 11.40 & \cellcolor{rankone}27.99 & 3.71 \\
    Qwen3.5-27B     & \cellcolor{rankone}34.53 & \cellcolor{rankone}20.59 & \cellcolor{rankone}10.93 & \cellcolor{ranktwo}29.58 & 2.87 \\
    InternVL3.5-38B & 51.17 & 34.68 & 23.12 & 44.75 & 6.10 \\
    \bottomrule
    \end{NiceTabular}}
  \end{minipage}
  \hfill %
  \begin{minipage}[c]{0.45\linewidth}
    \centering
    \includegraphics[width=\linewidth]{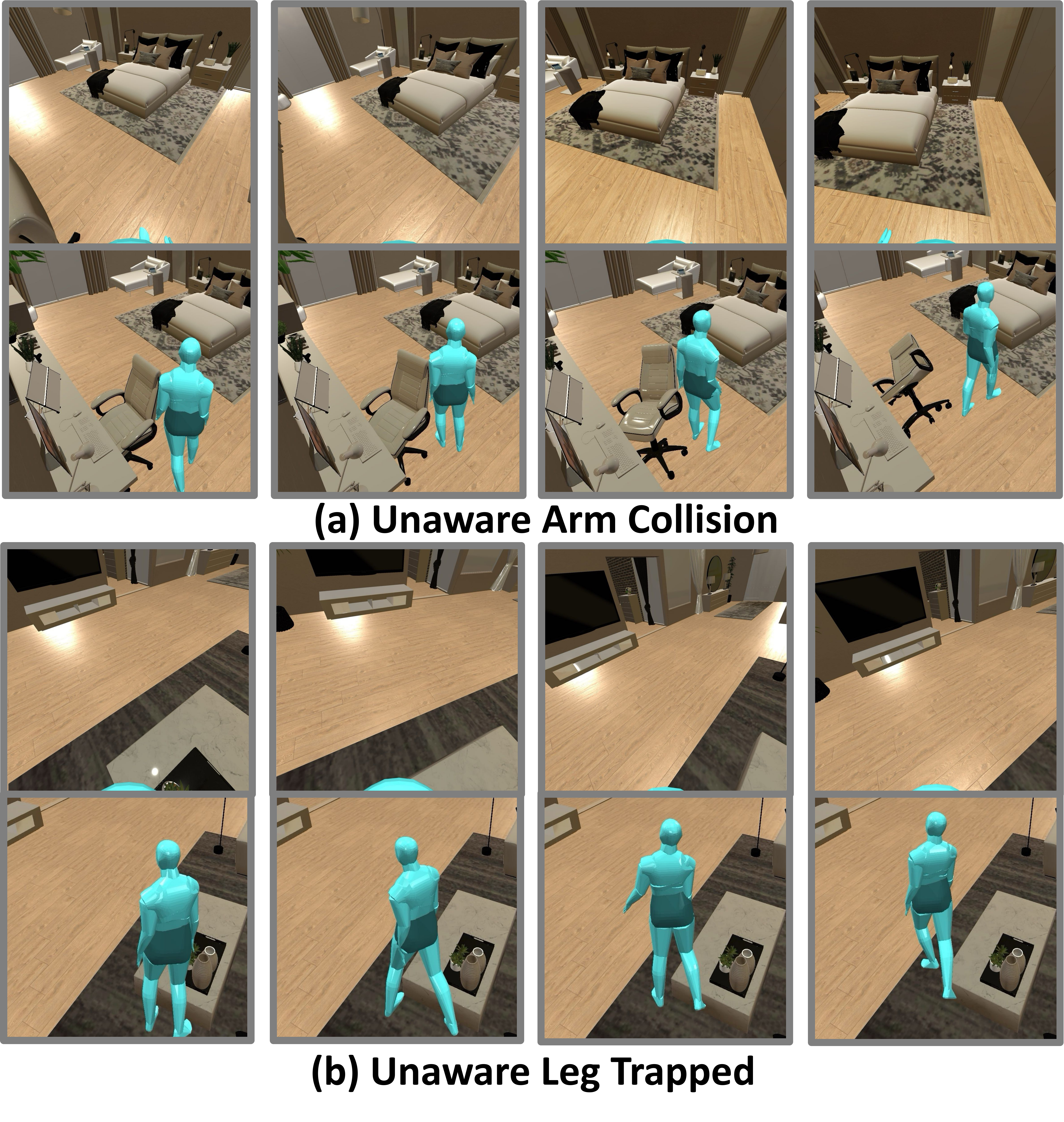}
    
    \captionof{figure}{\textbf{Body-unawareness failures.} Two episodes, each as
    egocentric (top) and third-person (bottom) views over consecutive steps:
    \textbf{(a)} the arm scrapes furniture unnoticed; \textbf{(b)} the legs jam against a
    low object while forward commands continue---the costs captured by \textbf{Coll.}\ and
    \textbf{\#Dtb}.}
    \label{fig:collision}
  \end{minipage}

\end{figure}

\begin{finding}{5}
Reaching is not interacting.
\end{finding}
Among the 726 episodes across the same nine VLMs in which the agent \emph{successfully navigates}
to the seat---it actively stops within $0.2$\,m of the target and thus already satisfies
navigation---513 (71\%) still fail to sit. We replay every such trajectory and, from the
pelvis--object mesh contacts logged along it (exact contact, not bounding-box overlap) together
with the action stream, attribute each failure to a single root cause (Fig.~\ref{fig:error_analysis}).
The failures are overwhelmingly errors of \emph{body awareness}---the agent commits to sitting
but puts its body in the wrong place (\textbf{81\%})---rather than errors of \emph{decision}
(\textbf{19\%}). The dominant mode is \textbf{sitting into thin air} (\textbf{58\%}): the agent
issues a sit, yet at no step does its pelvis touch any surface---it simply lowers its body where
it is standing, with no seat beneath it. Another \textbf{14\%} \emph{sit on the wrong thing}: the
pelvis lands on a different object or on the floor rather than the target seat (most often when
the goal is a couch, bed, or toilet, whose neighbours invite a mistaken landing). A final
\textbf{9\%} briefly \emph{do reach the seat}---the pelvis makes genuine mesh contact during a
sit---but the agent then keeps acting (predominantly turning, with occasional stands and
re-sits) instead of stopping, and shifts its body back off the seat before the last frame. The
remaining \textbf{19\%} are \emph{decision} failures: the agent reaches and stops standing in
front of the seat but never issues a sit at all. In short, once the humanoid has walked up to the
target, sitting is gated not by \emph{whether} it decides to sit but by \emph{where it puts its
body}: four in five failures drop the pelvis onto empty space, the wrong object, or off the
seat---a finer-grained instance of the same egocentric spatial deficit that limits navigation.
What interaction ultimately demands is thus the finest form of embodied spatial reasoning:
placing the body precisely with respect to both itself and the surrounding scene, and
\emph{anticipating the spatial relation each action will produce}---where the pelvis will
land relative to the seat once the sit unfolds---rather than merely deciding to act.

\begin{finding}{6}
No current VLM perceives its own body.
\end{finding}
Underlying failures at every stage is the agent's obliviousness to its own body.
Collisions concentrate in the parts it does not watch: across all nine models the legs
and feet collide most ($28$--$45\%$ of steps), the arms and hands next ($20$--$35\%$),
while the head---seldom near obstacles---almost never does (under $7\%$,
Table~\ref{tab:hcb_coll_bodygroup}). Two rollouts make the failure concrete
(Fig.~\ref{fig:collision}): in (a) the agent's arm knocks over a chair that was plainly
in view a few frames earlier, yet it never registers the contact; in (b) an obstacle
stays continuously visible straight ahead, yet the agent keeps walking its legs into it.
In both, the scene is seen but the body is not. A VLM trained purely by visual question
answering behaves like a \emph{ghost}: it holds no instinctive model of which pixels are
its own limbs, and it never tries to infer where they are. Part of this blindness is
inherited from the interface itself: the agent receives only egocentric RGB and text
history, with no proprioceptive body state or contact signal, so any sense of its own
body can only be reconstructed from vision alone (\Cref{sec:discussion}). This is not a peripheral
nuisance---body awareness underlies $34\%$ of navigation failures and $81\%$ of
interaction failures (Findings~4--5)---but a core capability that embodied action has
yet to acquire: \emph{feeling} the body, not merely seeing the world.

\definecolor{rankone}{RGB}{255,255,255}
\definecolor{ranktwo}{RGB}{255,255,255}

%% file: content/2_related.tex
\section{Related Work}
\label{sec:related}

\subsection{Embodied Tasks and Interactive Environments}
Embodied environments differ not only in the tasks they contain, but also in how an agent's actions are grounded and executed in the simulation.
VirtualHome~\citep{puig2018virtualhome} models household activities as programs over structured scene states and renders atomic instructions with predefined avatar animations, prioritizing activity semantics over closed-loop physical control. 
ALFRED~\citep{shridhar2020alfred} introduces egocentric language following, but exposes a discrete set of navigation and interaction actions whose execution is resolved by the simulator. 
Habitat~2.0~\citep{szot2021habitat2} and BEHAVIOR-1K~\citep{li2023behavior1k} move toward physics-enabled rearrangement and mobile manipulation, supporting articulated objects and persistent physical state under robot-specific control. 
Habitat~3.0 further introduces humanoid avatars and human-in-the-loop interaction, primarily to study robots acting around or collaborating with human partners~\citep{puig2024habitat3}.

Existing embodied environments are often mismatched to general-purpose MLLMs.
Program-driven environments resolve actions through predefined transitions, leaving little need to reason about body motion or physical consequence. 
Fully physics-based simulators, by contrast, require high-frequency continuous control and entangle spatial decisions with motor execution. 
HumanCLAW strikes a middle ground: a frozen MLLM selects parameterized whole-body motion skills based on egocentric observations, utilizing motion generation model autoregressively generate continuous long horizon behavior. 
Semi-physics execution preserves collision, contact, and object disturbance while abstracting low-level control, enabling the basic embodied interaction.

Closest to our find-and-navigate setting are object-goal navigation and
vision-and-language navigation, where an embodied agent locates and moves to a
target from egocentric observations~\citep{savva2019habitat,khanna2024hssd,anderson2018vln}.
These tasks, however, typically abstract the agent as a moving camera or a
velocity-controlled point and score only whether a location is reached, leaving out
how a full body arrives, orients, and interacts under physical consequence---exactly
the middle layer HumanCLAW targets.

\subsection{Multi-modal Foundation as Embodied Decision Makers}
Recent surveys distinguish hierarchical agentic systems from end-to-end vision--language--action (VLA) policies~\citep{ma2026vlasurvey,liang2025embodiedsurvey}. 
Hierarchical agents use multimodal large language models (MLLMs) for task decomposition, spatial grounding, skill selection, feedback-driven replanning~\citep{ichter2023saycan,liang2022code,huang2023innermonologue,song2023llmplanner,driess2023palme,huang2023voxposer,duan2024manipulate,li2026genhsi}, and open-world memory with reusable skills~\citep{wang2023deps,wang2024voyager}.
Frozen VLMs have also been driven in closed loop through first-person games~\citep{tan2024cradle,zhang2025videogamebench}, where actions remain key presses whose outcomes are resolved by the game script.
Although suited to long-horizon reasoning, they often reduce execution to symbolic actions or skill outcomes, obscuring body-level embodied interaction with environments.
End-to-end VLA policies instead learn direct motor grounding from embodied data~\citep{zitkovich2023rt2,ghosh2024octo,kim2025openvla,black2024pi0, bjorck2025gr00t}, but couple task reasoning to robot-specific data and low-level control.
Action-reasoning models such as MolmoAct~\citep{lee2025molmoact} externalize an editable mid-level spatial trace before emitting actions, yet the decision maker is still trained on action data and evaluated with execution in the loop, rather than queried as a frozen generalist.

Existing benchmarks assess goal interpretation, action sequencing, and long-horizon planning~\citep{li2024embodiedinterface,choi2024lotabench,chang2025partnr,yang2025embodiedbench,hong2026esibench}, yet primarily score plans, simulator actions, or robot trajectories. 
HumanCLAW instead evaluates how frozen off-the-shelf VLMs sustain egocentric spatial action over long-horizon tasks.
Their skill-level decisions are realized as continuous whole-body motion in a physical world, which preserves collision, contact, gravity, and object dynamics while largely abstracting away balance and motor-tracking failures. 
This setup evaluates progressive task success together with action coherence, body awareness, and environmental disturbance.

A growing line of work probes the spatial intelligence of VLMs
directly~\citep{chen2024spatialvlm,yang2024thinking,yang2025embodiedbench,li2025stare}, reporting
that current models struggle with metric distance, orientation, and multi-step
spatial simulation.
These evaluations, however, mostly pose spatial understanding as question
answering over images, videos, or cross-view pairs, in open-loop settings
without environmental feedback.
Action intelligence, as we use the term, is the operational component of
spatial intelligence that such settings leave unmeasured: converting spatial
understanding into closed-loop embodied decisions---selecting,
parameterizing, and sequencing actions as their physical consequences
unfold---with the model's own body state part of the problem.
It is likewise narrower than a planner in the generic sense (given an
observation and a goal, output what the agent should do next): a planner is
typically scored open-loop on the plausibility of its proposals, whereas
action intelligence is measured through execution, where every decision is
realized as motion and its consequence returns in the next egocentric view.
Our error analysis is consistent with the spatial-reasoning findings: even when
perception succeeds, the dominant failures are egocentric-spatial---self-localization,
reachability, and knowing when the body has arrived---so HumanCLAW serves as a
physically grounded, closed-loop stress test of exactly this bottleneck.

\subsection{Controllable Human Motion Generation as Agentic Tools}
Text-to-motion models learn strong priors for plausible full-body movement from large motion-language corpora~\citep{guo2022humanml3d,tevet2023mdm,guo2024momask,huang2024controllable}. 
Beyond just using text, following controllable motion generation methods further condition motion on spatial constraints, character identity, human interaction, ego perception, physics, and music~\citep{hassan2021samp,zhang2022gamma,zhao2023dimos,xie2024omnicontrol,yuan2023physdiff,siyao2025half,wu2025text2interact,jia2026iam,cong2026umo,zhang2025egoreact,Li_2026_CVPR,zhang2025egoreact,diomataris2025moving,zhang2025primal,siyao2022bailando,siyao2023bailandopp,siyao2024duolando}. 
Although additional control modalities improve motion fidelity, their input protocols beyond the output format of existing MLLMs agent, i.e., text-only output. 
An agentic executable tool set requires a compact and predictable interface: each tool-calling should follow its semantic intent and continuous parameters while producing a reliable local outcome. 
Therefore, HumanCLAW proposes a shared receding-horizon motion prior and a plug-and-play adapter for each atomic skill based on text templates with parameters, leaving long-horizon skill selection, parameterization, and composition to VLMs.

%% file: content/6_discussion.tex
\section{Discussion}
\label{sec:discussion}

HumanCLAW isolates a middle layer of physical action---continuous whole-body motion
with real scene consequences, but without balance or motor-tracking failures---and uses
it to ask how well today's VLMs act through a body. Across nine state-of-the-art models
the answer is consistent: none solves the benchmark, and the failures do not lie where a
vision--language agent is usually judged. Perception is largely intact---once a target
is rendered it is almost always recognized (Finding~3)---yet competence collapses in
everything that follows recognition. What is missing is \emph{embodied
self-awareness}: knowing where the body is and when it has arrived (Finding~4), placing
it precisely enough to interact (Finding~5), and sensing when it collides with the world
(Finding~6). The common thread is that current VLMs reason about the scene but not about
the body they now inhabit.

We attribute this to how these models are built. A VLM trained by visual question
answering learns to describe what it sees, not to feel what it does; it treats its own
limbs as just more pixels and never forms the proprioceptive, consequence-predicting
model that acting through a body requires. The result behaves like a ghost---fluent
about the world, oblivious to itself. Closing this gap will likely need more than
sharper recognition or a longer context window: it calls for persistent spatial memory,
calibrated termination, and an internal model of the body together with the spatial
relations each action produces.

Beneath the framework lies a philosophical commitment: we hold that generalizable
action intelligence arises from reasoning, not from fitting action data. A policy
fitted on trajectories generalizes within the support of its data; a reasoner
generalizes as far as its knowledge extrapolates. HumanCLAW is built on this
bet---the decision maker stays frozen and general, the motor layer is a fixed,
reusable prior, and all task-level composition happens zero-shot in reasoning. On
this view, a general action agent improves not by collecting ever more trajectories
but by becoming a better reasoner: new skills register without retraining, new tasks
require only new rules, and every advance in foundation models transfers to the body
for free. Our results leave the bet open but sharpened: today's reasoning does not
yet reach the body, yet the failures are decision-level---exactly the layer where
better reasoning, rather than more data, could be brought to bear. Absent a
comparison against fitted action policies, this last step remains our working
hypothesis: what the experiments establish is where the failures sit, not which
route will close them.

Our study has limits. The interaction vocabulary is small;
richer manipulation would broaden the interact stage and stress finer body placement.
Decision-level attribution is likewise relative to the skill vocabulary: with
motor execution factored out, what reads as a decision failure still depends
on the granularity of the atomic skills, and a finer or coarser vocabulary
would draw the boundary elsewhere.
Half-physics deliberately abstracts away balance and motor tracking, so HumanCLAW
measures action decisions rather than low-level control, and transferring a competent
decision maker onto a physical humanoid remains future work.
The interface likewise simulates no tactile channel: collisions displace the world
but are never felt, so the agent can achieve proprioception only through its
egocentric view---which may be intrinsically difficult, and a body-state or contact
signal could be the missing input rather than a missing faculty. Finally, we evaluate frozen
off-the-shelf VLMs; whether targeted training for body awareness can lift these ceilings
is an open---and, we believe, tractable---question.

HumanCLAW-Bench leaves ample headroom: the strongest model completes the full
task on only $16.8\%$ of episodes, and even navigation alone peaks at
$42.4\%$---headroom concentrated exactly where progress in embodied self-awareness will
compound. We offer it as a clean testbed for the next generation of agents---agents
that not only see and plan, but act.

%% file: content/appendix.tex
\section{Skill-Specific Verifier Checks}
\label{app:verifier}

The verifier applies a fixed, skill-triggered checklist. Its purpose is to
counter long-context degradation: as the rollout history grows, the planner's
estimates of spatial distance and relation become unreliable, so each check is
re-answered from the current egocentric view alone, through a compact,
skill-specific prompt. The verifier runs only when one of the skills in
\Cref{tab:app_verifier} is proposed; if a check fails, the proposal is replaced
with a corrected skill from the same action pool. Skills whose precondition
cannot be assessed from the egocentric view---e.g.\ stepping back, where the
region behind the body is unobservable---carry no check.

\begin{table}[h]
  \centering
  \small
  \setlength{\tabcolsep}{5pt}
  \renewcommand{\arraystretch}{1.25}
  \begin{tabular}{@{}p{0.23\linewidth}p{0.72\linewidth}@{}}
    \toprule
    Trigger skill & Checks \\
    \midrule
    walk forward & Is the path ahead clear; would the motion collide with a wall
    or furniture; does moving forward advance the current goal. \\
    stop & Is the high-level goal actually complete; has the body arrived at, or
    made contact with, the target. \\
    climb up & Is a stair step directly underfoot; is the body facing the
    staircase. \\
    climb down & Same checks as climb up: a step directly underfoot and the body
    facing the staircase. \\
    turn (before sitting) & Is the body already flush against the seat (zero
    distance); after the turn, would the seat lie directly behind the body. \\
    \bottomrule
  \end{tabular}
  \caption{\textbf{Skill-specific verifier checks.} Each rule fires only when its
  skill is proposed; a failed check replaces the proposal with a corrected skill
  from the same action pool.}
  \label{tab:app_verifier}
\end{table}

\section{Automated Root-Cause Attribution}
\label{app:rootcause}

All root-cause labels in \Cref{fig:error_analysis}(b--d) are assigned
automatically---no human annotation is involved. A deterministic classifier
replays each failed episode's logs and evaluates, per decision step, the
geodesic distance from the pelvis to the nearest target instance on the scene's
navigation mesh, pelvis--object mesh contacts, the target's rendered semantic
pixel count in the egocentric view, the action stream, and the model's own
stated visible state. Within each stage, the rules in
\Cref{tab:app_rootcause} are evaluated top to bottom and the first matching
rule assigns the label, so every failed episode receives exactly one error
type and the labeling is exactly reproducible from the logs.

\begin{table}[h]
  \centering
  \small
  \setlength{\tabcolsep}{5pt}
  \renewcommand{\arraystretch}{1.25}
  \begin{tabular}{@{}p{0.23\linewidth}p{0.72\linewidth}@{}}
    \toprule
    Error type & Rule (evaluated top to bottom; first match assigns the label) \\
    \midrule
    \multicolumn{2}{@{}p{0.96\linewidth}@{}}{\textbf{Find} failures --- no step both
    renders the target with $\geq$$100$ semantic pixels in the $512\times512$
    egocentric view and has the model acknowledge it} \\[1pt]
    target in view but not seen & The target renders $\geq$$100$ pixels at some step,
    but the model never acknowledges it in its visible-state output. \\
    near but not in view & The pelvis comes within $1.5$\,m of a target instance, yet
    the target never renders $\geq$$100$ pixels: the agent reaches the object without
    ever bringing it into view. \\
    too far & The geodesic distance from the starting pose to the nearest target
    instance exceeds $20$\,m. \\
    unaware jammed & The agent issues locomotion commands for $\geq$$8$ consecutive
    steps while per-step body displacement stays below $0.1$\,m. \\
    ineffective actions & A window of $\geq$$25$ consecutive steps in which the agent
    keeps issuing movement or turning commands while the body remains confined within
    a $2$\,m box. \\
    midpoint give up & The agent first cuts its geodesic distance to the target to
    below half the initial value, but then retreats until the episode ends at more
    than $0.6\times$ the initial distance. \\
    ineffective exploration & All remaining cases in which the agent never brings
    the target into view. \\
    \midrule
    \multicolumn{2}{@{}p{0.96\linewidth}@{}}{\textbf{Nav} failures --- the target is
    actively found, but the agent never both actively stops and brings the pelvis
    within $0.2$\,m of the target} \\[1pt]
    unaware arrived/touch & The pelvis comes within $0.2$\,m of the target at some
    step, but the agent never actively stops: it keeps issuing navigation actions
    after arrival. \\
    stop while far & The agent actively stops and declares arrival, but the pelvis
    has never come within $0.2$\,m of the target. \\
    find too late & The target is first rendered only after step $70$ of the
    $100$-step budget, leaving too few steps to reach it. \\
    unaware jammed & The target is still in view ($\geq$$100$ pixels) at the final
    frame, and the trajectory contains $\geq$$8$ consecutive locomotion steps with
    body displacement below $0.1$\,m: the agent is lodged against geometry yet keeps
    issuing movement commands. \\
    ineffective actions & The target is in view at the final frame and the body moves
    freely (no jammed run), yet the agent's actions never bring it within
    $0.2$\,m. \\
    target hallucination & The target renders $<$$100$ pixels at the final frame, but
    the model's final visible-state still claims to see it. \\
    lost target & The target is out of view at the final frame and the model no
    longer claims to see it: the agent turned away and never re-acquired it. \\
    \midrule
    \multicolumn{2}{@{}p{0.96\linewidth}@{}}{\textbf{Interact} failures --- navigation
    succeeds (active stop within $0.2$\,m), but the sit fails} \\[1pt]
    stand after sit & Some sit step achieves pelvis--seat mesh contact with the
    target, yet the agent continues acting and its pelvis is off the seat at the
    final frame. \\
    sit on air & A sit is issued, but the pelvis never makes mesh contact with any
    surface throughout the episode. \\
    sit wrong & A sit is issued and the pelvis ends resting on a different object,
    the floor, or a wall rather than the target seat. \\
    stop but never sit & The agent reaches the seat and stops, but never issues a
    sit. \\
    \bottomrule
  \end{tabular}
  \caption{\textbf{Automated root-cause rules} behind the per-stage breakdowns in
  \Cref{fig:error_analysis}(b--d). All quantities are read from the logged
  rollouts (navmesh geodesic distances, pelvis--object mesh contacts, per-step
  egocentric renders, the action stream, and the model's stated visible state);
  contact is exact mesh contact, not bounding-box overlap.}
  \label{tab:app_rootcause}
\end{table}